\documentclass[10pt,twocolumn,letterpaper]{article}

\usepackage{cvpr}
\usepackage{wrapfig}
\usepackage{placeins}
\usepackage{multirow}
\usepackage{lineno}
\usepackage{booktabs}
\usepackage{float}

\definecolor{cvprblue}{rgb}{0.21,0.49,0.74}
\usepackage[pagebackref,breaklinks,colorlinks,allcolors=cvprblue]{hyperref}

\def\confName{CVPR}
\def\confYear{2025}

\title{DCT-Mamba3D: Spectral Decorrelation and Spatial-Spectral Feature Extraction for Hyperspectral Image Classification}
\author{
    Weijia Cao$^{1}$ \quad
    Xiaofei Yang$^{2}$ \quad
    Yicong Zhou$^{3}$ \quad
    Zheng Zhang$^{4}$ \\[3pt]
    $^1$Aerospace Information Research Institute, Chinese Academy of Sciences, Beijing, China \\
    $^2$Guangzhou University, Guangzhou, China \\
    $^3$University of Macau, Macau, China \\
    $^4$Harbin Institute of Technology, Shenzhen, China \\[3pt]
    {\tt\small caowj@aircas.ac.cn, xiaofeiyang@gzhu.edu.cn, yicongzhou@um.edu.mo, darrenzz219@gmail.com}
}
\begin{document}

% Title and Abstract
\maketitle
\begin{abstract}
Hyperspectral image classification presents challenges due to spectral redundancy and complex spatial-spectral dependencies. This paper proposes a novel framework, DCT-Mamba3D, for hyperspectral image classification. DCT-Mamba3D incorporates: (1) a 3D spectral-spatial decorrelation module that applies 3D discrete cosine transform basis functions to reduce both spectral and spatial redundancy, enhancing feature clarity across dimensions; (2) a 3D-Mamba module that leverages a bidirectional state-space model to capture intricate spatial-spectral dependencies; and (3) a global residual enhancement module that stabilizes feature representation, improving robustness and convergence. Extensive experiments on benchmark datasets show that our DCT-Mamba3D outperforms the state-of-the-art methods in challenging scenarios such as the same object in different spectra and different objects in the same spectra.
\end{abstract}

% Main Sections
\section{Introduction}

\begin{figure}[H]
    \centering
    \begin{subfigure}[b]{1\linewidth}
        \includegraphics[width=\linewidth]{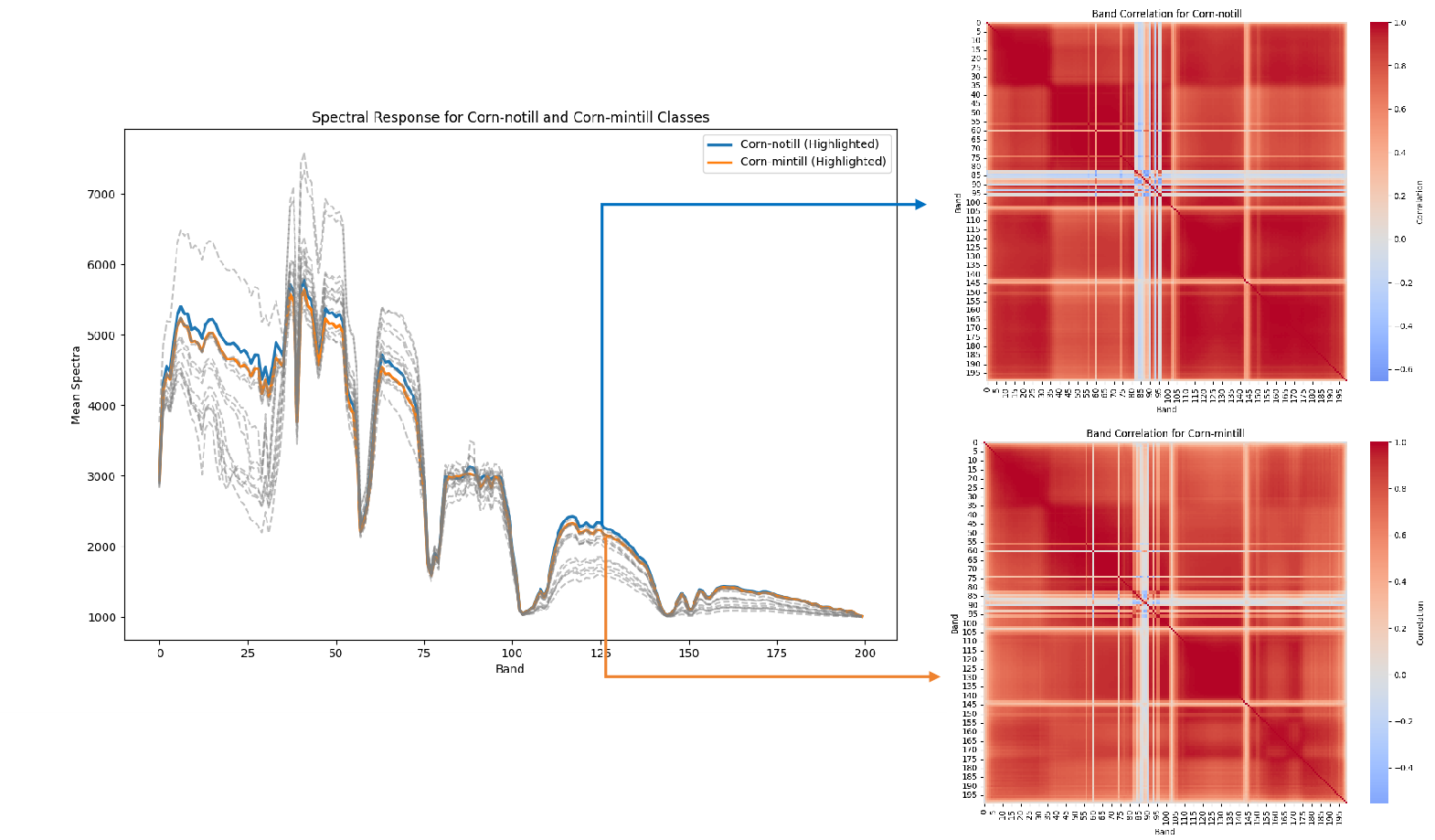}
        \caption{Same object, different spectra: Spectral variability in corn types (\textit{Corn-notill} and \textit{Corn-mintill}). Highlighted curves represent specific types, while gray curves show other corn samples, reflecting intra-class variability influenced by spectral redundancy and high correlation.}
    \end{subfigure}
    \vfill
    \begin{subfigure}[b]{1\linewidth}
        \includegraphics[width=\linewidth]{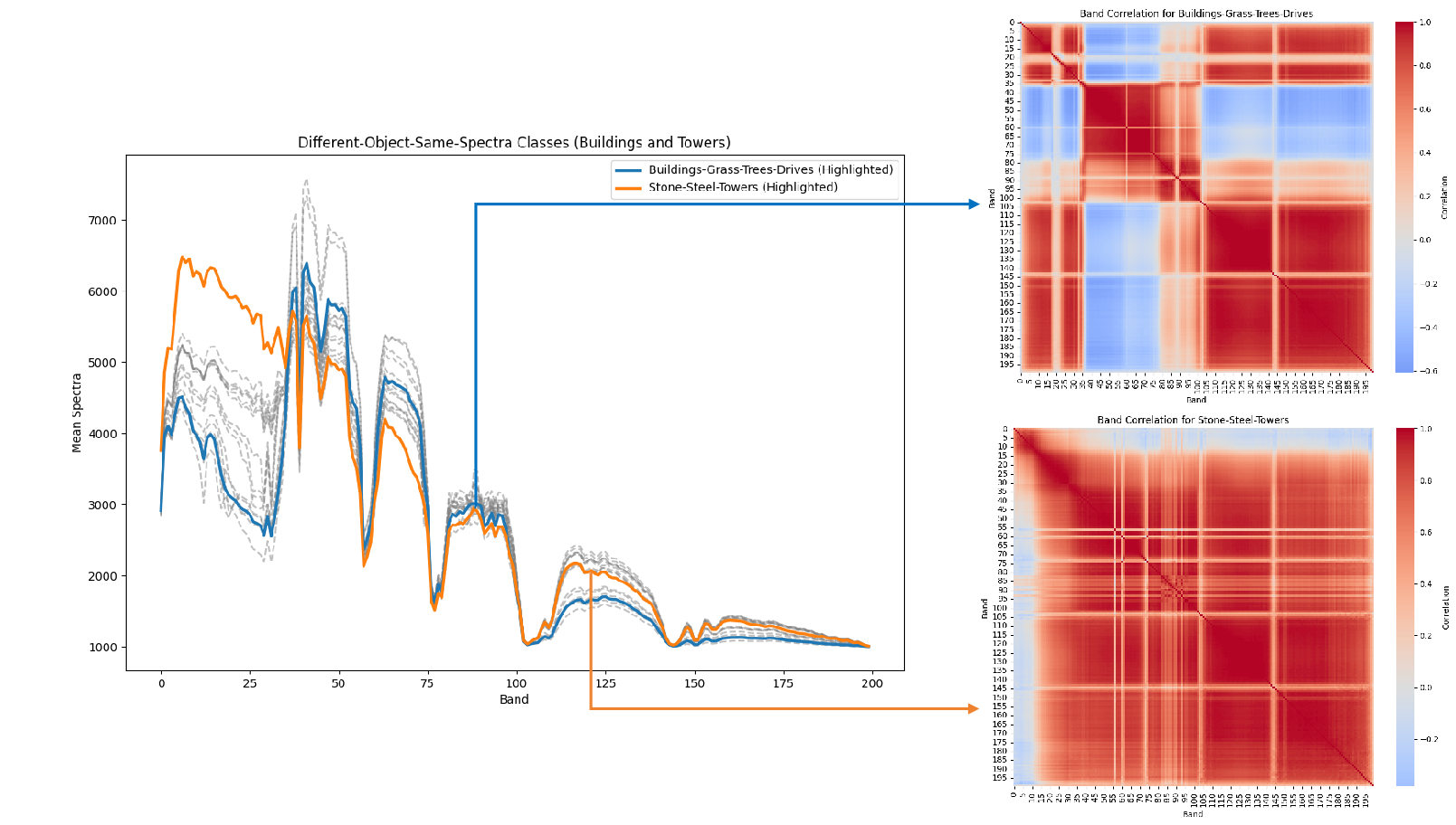}
        \caption{Different objects, same spectra: Spectral similarity between \textit{Buildings-Grass-Trees-Drives} and \textit{Stone-Steel-Towers}. Highlighted curves represent these classes, with gray curves showing other land cover types, illustrating inter-class similarity caused by spectral overlap and strong correlation.}
    \end{subfigure}
    \caption{Spectral response functions illustrating HSI classification challenges due to spectral redundancy and high correlation.}
    \label{fig:challenges}
\end{figure}

Hyperspectral image (HSI) classification is crucial in remote sensing applications, such as environmental monitoring, agriculture, and mineral exploration~\cite{zhang2016deep, paoletti2019deep}. However, high dimensionality and spectral redundancy in HSI data—often termed the “curse of dimensionality”—pose unique challenges, complicating effective classification~\cite{he2017recent, li2013spectral}. This redundancy hampers performance, particularly in cases where different objects share similar spectra or the same object exhibits spectral variability under different conditions~\cite{theiler2019spectral, yao2024spectralmamba}. Figure~\ref{fig:challenges} illustrates these phenomena, emphasizing the need for methods that capture essential spatial-spectral features while reducing redundant information~\cite{li2024latent,deng2023psrt}.

\begin{figure*}[t]
    \centering
    \includegraphics[width=1\linewidth]{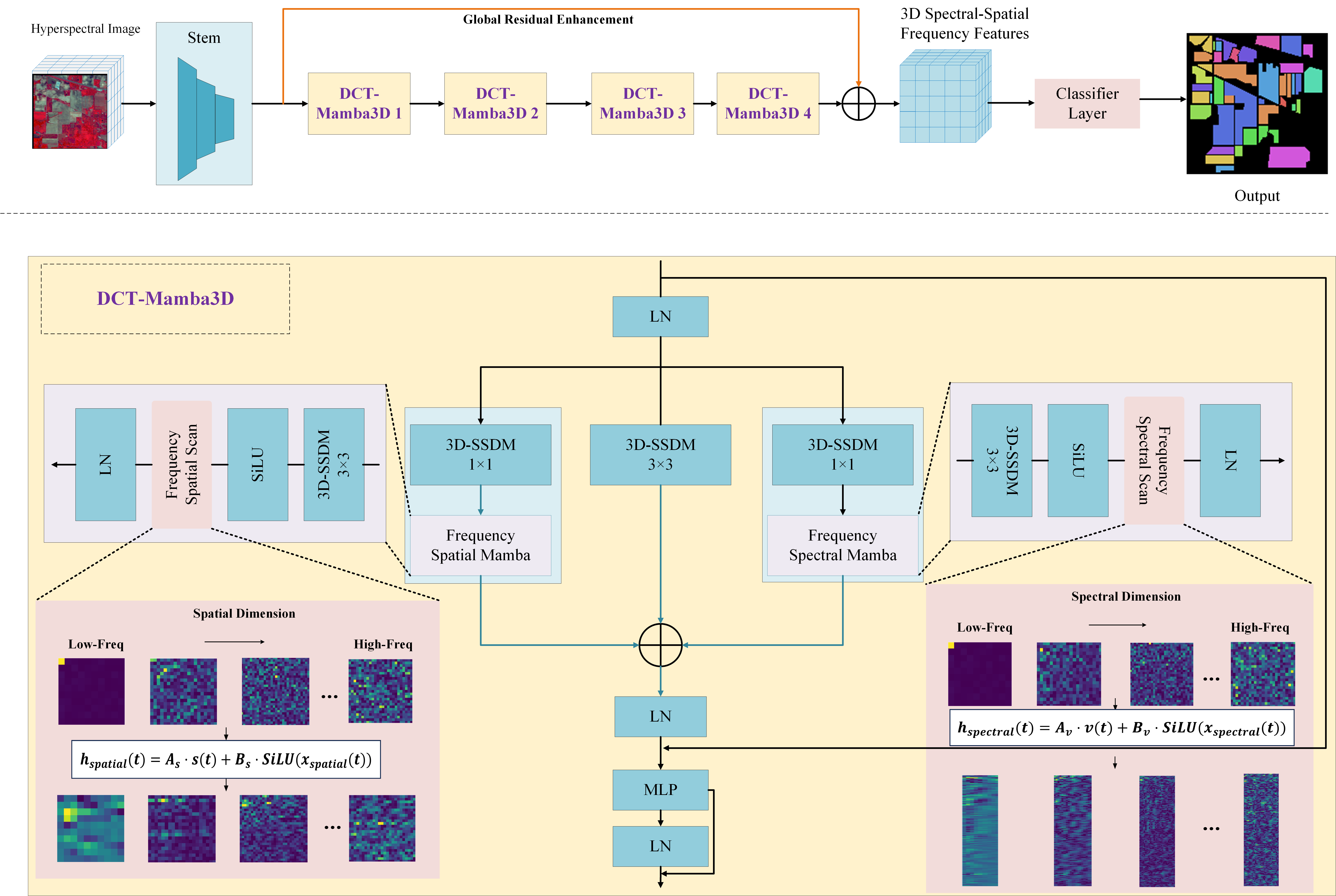}
    \caption{DCT-Mamba3D framework.}
    \label{fig:dct-mamba3d-flowchart}
\end{figure*}

\textbf{Challenges of Spectral Variability and Redundancy}: Spectral variability (caused by changing illumination, atmospheric conditions, or intrinsic material differences) and similarity between materials exacerbate classification challenges~\cite{theiler2019spectral, yao2024spectralmamba, hong2024spectralgpt}. High inter-band correlation leads to redundant information, complicating differentiation, particularly in mixed pixels where each pixel may represent multiple materials~\cite{be2014tgrs}.

\textbf{Frequency-Domain Transformations for Enhanced Feature Extraction}: Frequency-domain transformations can improve spectral separation and feature extraction in HSI classification~\cite{yan2024exploiting, qiao2023dual}. Discrete cosine transform (DCT) specifically enables decorrelation by transforming data into the frequency domain, facilitating refined feature extraction~\cite{shen2021dct, ulicny2022harmonic, xu2020learning, ulicny2019harmonic}.

Recent approaches in HSI classification explore CNN-based, Transformer-based, and Mamba-based architectures. CNN-based methods, such as 2D-CNN~\cite{yang2018hyperspectral}, 3D-CNN~\cite{yang2020synergistic}, and HybridSN~\cite{roy2019hybridsn}, primarily focus on spatial features but often overlook complex spectral correlations~\cite{chen2016deep, Jia2022tgrs, ulicny2022harmonic, Xu2023grsl}. Transformer-based models, including ViT~\cite{dosovitskiy2020image}, HiT~\cite{yang2022hyperspectral}, CAT~\cite{feng2024cat}, and MorphF~\cite{roy2023spectral}, utilize self-attention mechanisms to capture spectral dependencies but are computationally intensive and often require large datasets~\cite{scheibenreif2023masked, feng2024cat}. Mamba-based models, such as MiM~\cite{zhou2024mamba}, SpectralMamba~\cite{yao2024spectralmamba}, WaveMamba~\cite{ahmad2024wavemamba}, and Vision Mamba~\cite{zhu2024vision}, employ state-space representations to model spatial-spectral relationships without convolutional structures but face limitations in addressing spectral redundancy and inter-band correlation.

In this paper, we propose \textit{DCT-Mamba3D}, an HSI classification model that integrates a 3D Spatial-Spectral Decorrelation Module (3D-SSDM), a 3D-Mamba module, and a Global Residual Enhancement (GRE) module to reduce spectral redundancy and enhance feature extraction. 3D-SSDM uses 3D DCT basis functions to transform data into the frequency domain, enabling both spectral and spatial decorrelation and improving feature clarity for subsequent extraction layers. The 3D-Mamba module leverages a 3D state-space model to capture intricate spatial-spectral dependencies. Finally, the GRE module stabilizes feature representation, enhancing robustness and convergence.

Our contributions are as follows:
\begin{itemize}
    \item \textbf{Spectral-Spatial Decorrelation with 3D-SSDM}: The 3D Spatial-Spectral Decorrelation Module (3D-SSDM) reduces spectral and spatial redundancy using 3D DCT basis functions, enabling comprehensive feature separability in complex HSI scenarios.
    \item \textbf{Efficient Spatial-Spectral Dependency Modeling with 3D-Mamba}: The 3D-Mamba module captures both local and global spatial-spectral dependencies, enhancing efficiency and feature interaction.
    \item \textbf{Robust Feature Stability with GRE}: The Global Residual Enhancement (GRE) module stabilizes feature representation, improving robustness and convergence.
\end{itemize}

The rest of this paper is organized as follows: Section~\ref{sec:related_work} reviews related work, Section~\ref{sec:methodology} details our approach, Section~\ref{sec:experiments} presents results, Section~\ref{sec:discussion} discusses model advantages, and Section~\ref{sec:conclusion} concludes the paper.

\section{Related Work}
\label{sec:related_work}

This section reviews key approaches in hyperspectral image (HSI) classification, focusing on frequency-domain analysis, Transformer-based models, and Mamba-based models for capturing spatial-spectral dependencies.

\subsection{Frequency-Domain Techniques for Image Classification}

Frequency-domain techniques, including the discrete cosine transform (DCT)~\cite{shen2021dct, ulicny2022harmonic, xu2020learning, ulicny2019harmonic}, discrete Fourier transform (DFT)~\cite{zhang2024three, wang2019frequency}, and wavelet transformations~\cite{yao2022wave}, have proven effective in enhancing feature extraction for image classification. Techniques such as Harmonic Neural Networks (HNN)~\cite{ulicny2022harmonic, ulicny2019harmonic} apply 2D DCT to capture subtle variations in the frequency domain, exploiting DCT’s ability to decorrelate and concentrate energy, thereby reducing redundancy and improving classification performance in natural image classification~\cite{ahmed1974discrete}. Similarly, 2D FFT-based methods~\cite{qiao2023dual, zhang2024three, wang2019frequency} leverage frequency information to extract discriminative features for HSI classification. WaveViT~\cite{yao2022wave} utilizes discrete wavelet transformations for multi-scale feature extraction.

\subsection{Mamba-Based HSI Classification Approaches}

Mamba-based models present a highly promising approach for hyperspectral image (HSI) classification, focusing on capturing spatial-spectral dependencies through state-space representations. These models are particularly well-suited to the high-dimensional nature of HSI data, enabling effective feature extraction without relying on traditional convolutional or attention mechanisms. Notable Mamba-based methods, such as MiM~\cite{zhou2024mamba}, SpectralMamba~\cite{yao2024spectralmamba}, and WaveMamba~\cite{ahmad2024wavemamba}, leverage state-space models to capture spatial-spectral relationships, demonstrating the potential of Mamba for HSI classification. Recent advancements, including Li et al.~\cite{li2024mambahsi}, emphasize the importance of integrating spatial and spectral features. Despite their success in modeling long-range dependencies, these approaches still face significant challenges related to the redundancy between spectral bands, which remains a critical gap in current Mamba-based methods.

\section{Methodology} \label{sec:methodology}

Our proposed \textbf{DCT-Mamba3D} comprises three main components, as shown in Fig.~\ref{fig:dct-mamba3d-flowchart}. First,  \textbf{3D Spatial-Spectral Decorrelation Module (3D-SSDM)} applies 3D DCT basis functions to convert spatial pixels into decorrelated frequency components, reducing redundancy and isolating essential features. Second, the \textbf{3D-Mamba module} uses state-space modeling and selective scanning to capture complex spatial-spectral dependencies within the decorrelated data. Finally, the \textbf{GRE module} stabilizes feature representation across layers by integrating global context, enhancing robustness and classification accuracy.

\subsection{3D-SSDM Module}

\begin{figure}[t]
    \centering
    \includegraphics[width=1\linewidth]{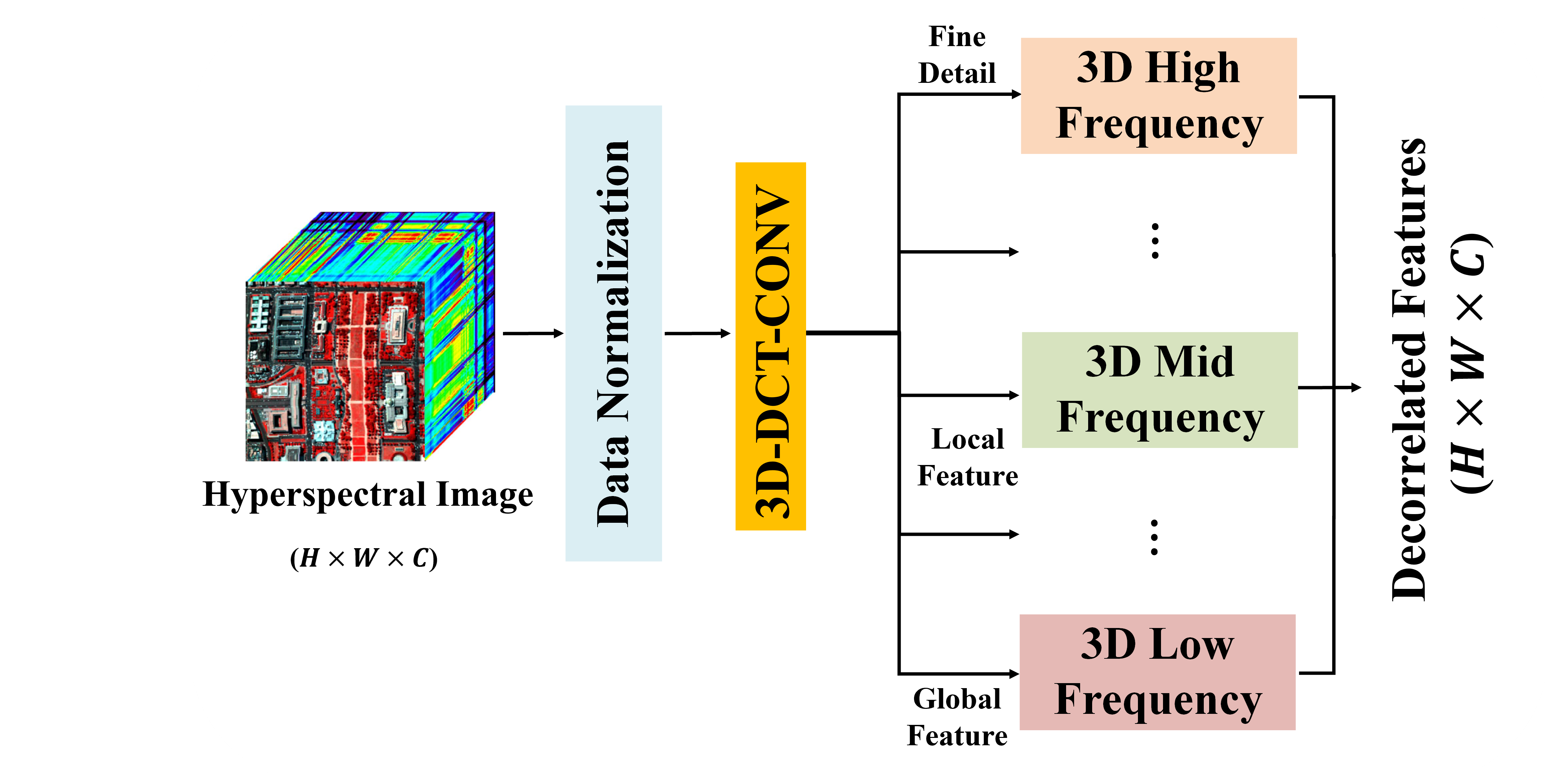}
    \caption{3D Spatial-Spectral Decorrelation Module (3D-SSDM), applying 3D DCT basis functions to decompose the image into independent frequency components, aiding decorrelation and feature extraction.}
    \label{fig:3dssdm}
\end{figure}

The \textbf{3D-SSDM Module} begins with a \textbf{Stem} stage for shallow feature extraction and normalization. HSIs contain hundreds of contiguous spectral bands with high inter-band correlation. To address this, 3D-SSDM applies \textbf{3D DCT basis functions} to convert spatial pixels into decorrelated frequency components across both spatial and spectral dimensions. It enhances feature clarity by consolidating most of the energy into distinct frequency components, as shown in Fig.~\ref{fig:3dssdm}.

The 3D DCT generates a set of spatial-spectral frequency components, capturing varying HSI characteristics. In a \(3 \times 3 \times 3\) setup, the 3D DCT yields 27 basis functions, with low frequencies capturing smooth variations and high frequencies capturing fine-grained details.

Representing the HSI as \( X \in \mathbb{R}^{H \times W \times C} \) (where \( H \), \( W \), and \( C \) are spatial and spectral dimensions), the 3D DCT is applied as:
\begin{equation}
X_{\text{freq}}(i, j, k) = \sum_{x=0}^{H-1} \sum_{y=0}^{W-1} \sum_{z=0}^{C-1} X(x, y, z) \cdot \psi_{i,j,k}(x, y, z),
\end{equation}
where \( \psi_{i,j,k}(x, y, z) \) represents the 3D DCT basis functions, decorrelating both spatial and spectral dimensions and extracting spatial-spectral frequency features.

The basis functions \( \psi_{i,j,k}(x, y, z) \) are defined as:
\begin{equation}
\begin{aligned}
&\psi_{i,j,k}(x, y, z) = \alpha_i \alpha_j \alpha_k \cos\left(\frac{\pi (2x + 1) i}{2H}\right) \\
& \times \cos\left(\frac{\pi (2y + 1) j}{2W}\right) \cos\left(\frac{\pi (2z + 1) k}{2C}\right),
\end{aligned}
\end{equation}
where normalization factors \( \alpha_n \) are:
\[
\alpha_n =
\begin{cases}
\sqrt{\frac{1}{N}} & \text{if } n = 0, \\
\sqrt{\frac{2}{N}} & \text{if } n > 0,
\end{cases}
\]
with \( N \) as \( H \), \( W \), or \( C \).

\subsection{3D-Mamba Module}

The \textbf{3D-Mamba module} employs a bidirectional state-space model (SSM) to capture spatial-spectral dependencies within the frequency-domain data \( X_{\text{freq}} \) from the 3D-SSDM. It operates directly on decorrelated data, distinguishing similar features and refining spatial-spectral information across all dimensions.

After decorrelation, the input \( X_{\text{freq}} \) undergoes the following stages:

\begin{itemize}
    \item \textbf{Patch Embeddings:} The 3D-Mamba module decomposes \( X_{\text{freq}} \) into spatial, spectral, and residual components via specialized embedding layers in the frequency domain. This setup separates essential features and prepares data for selective scanning:
    \begin{equation}
    \begin{aligned}
    x_{\text{spatial}} &= F_{\text{PE}_1}(X_{\text{freq}}), \quad x_{\text{spectral}} = F_{\text{PE}_2}(X_{\text{freq}}), \\
    x_{\text{residual}} &= F_{\text{PE}_3}(X_{\text{freq}}).
    \end{aligned}
    \end{equation}

    \item \textbf{Frequency Spatial and Spectral bidirectional SSM:} Using SiLU activation for non-linearity, \( x_{\text{spatial}} \) and \( x_{\text{spectral}} \) undergo selective scanning within the SSM framework, capturing spatial-spectral dependencies in independent frequency components:
        \begin{equation}
        \begin{aligned}
            h_{\text{spatial}}(t) &= A_s \cdot s(t) + B_s \cdot \text{SiLU}(x_{\text{spatial}}(t)), \\
            h_{\text{spectral}}(t) &= A_v \cdot v(t) + B_v \cdot \text{SiLU}(x_{\text{spectral}}(t)),
        \end{aligned}
        \end{equation}
        where \( s(t) \) and \( v(t) \) denote latent spatial and spectral states, and \( h_{\text{spatial}}(t) \), \( h_{\text{spectral}}(t) \) are the outputs, enhancing independence across frequency domains.

    \item \textbf{Feature Aggregation and Normalization:} The outputs \( h_{\text{spatial}}(t) \) and \( h_{\text{spectral}}(t) \) are combined with residuals \( x_{\text{residual}} \), and normalized for stability:
        \begin{equation}
        y_{\text{mamba}}(t) = \gamma_0 \cdot x_{\text{residual}} + \gamma_1 \cdot h_{\text{spatial}}(t) + \gamma_2 \cdot h_{\text{spectral}}(t),
        \end{equation}
        where \( y_{\text{mamba}}(t) \) is the final spatial-spectral feature map, combining initial decorrelated features with refined updates.
\end{itemize}

\subsection{GRE Module}

The \textbf{GRE module} enhances feature robustness by integrating global context with spatial-spectral features extracted by the 3D-Mamba module. By introducing a residual connection, the GRE module stabilizes training and preserves key information across layers.

The GRE module receives \( y_{\text{mamba}} \) from the 3D-Mamba module and combines it with \( X_{\text{freq}} \) from the 3D-SSDM to form the final output \( F_{\text{out}} \):
\begin{equation}
F_{\text{out}} = y_{\text{mamba}} + \alpha X_{\text{freq}},
\end{equation}
where \( F_{\text{out}} \) is the final feature map, and \( \alpha \) is a learnable parameter balancing contributions from \( y_{\text{mamba}} \) and \( X_{\text{freq}} \).

The output \( F_{\text{out}} \) feeds into the classification layer, completing the feature extraction pipeline for HSI classification.

For optimization, a composite loss function is used, combining cross-entropy loss and an optional regularization term for spectral decorrelation:
\begin{equation}
\mathcal{L} = \mathcal{L}_{\text{CE}} + \lambda \mathcal{L}_{\text{reg}},
\end{equation}
where \( \mathcal{L}_{\text{CE}} \) is the cross-entropy loss, \( \mathcal{L}_{\text{reg}} \) penalizes spectral redundancy, and \( \lambda \) is a regularization weight.

\section{Experiments}
\label{sec:experiments}

\subsection{Datasets and Setup}
We evaluate DCT-Mamba3D on three benchmark hyperspectral image (HSI) classification datasets: Indian Pines, Kennedy Space Center (KSC), and Houston2013, each containing diverse land cover classes with distinct spectral characteristics. The Indian Pines dataset, featuring challenges like "same object, different spectra" and "different objects, same spectra," is especially suited for analyzing our model's spectral decorrelation capabilities. All datasets are split with 10\% for training and 90\% for testing, and model performance is averaged over 10 runs for robustness. Evaluation metrics include overall accuracy (OA), average accuracy (AA), Kappa coefficient, and F1-score per class. Our method is compared with leading models, including 2D-CNN~\cite{yang2018hyperspectral}, 3D-CNN~\cite{yang2020synergistic}, HybridSN~\cite{roy2019hybridsn}, ViT~\cite{dosovitskiy2020image}, HiT~\cite{yang2022hyperspectral}, MorphF~\cite{roy2023spectral}, SSFTT~\cite{sun2022spectral}, and MiM~\cite{zhou2024mamba}.

\subsection{Spectral Correlation Heatmaps}
\begin{figure}[H]
    \centering
    \begin{subfigure}[b]{0.32\linewidth}
        \includegraphics[width=\linewidth]{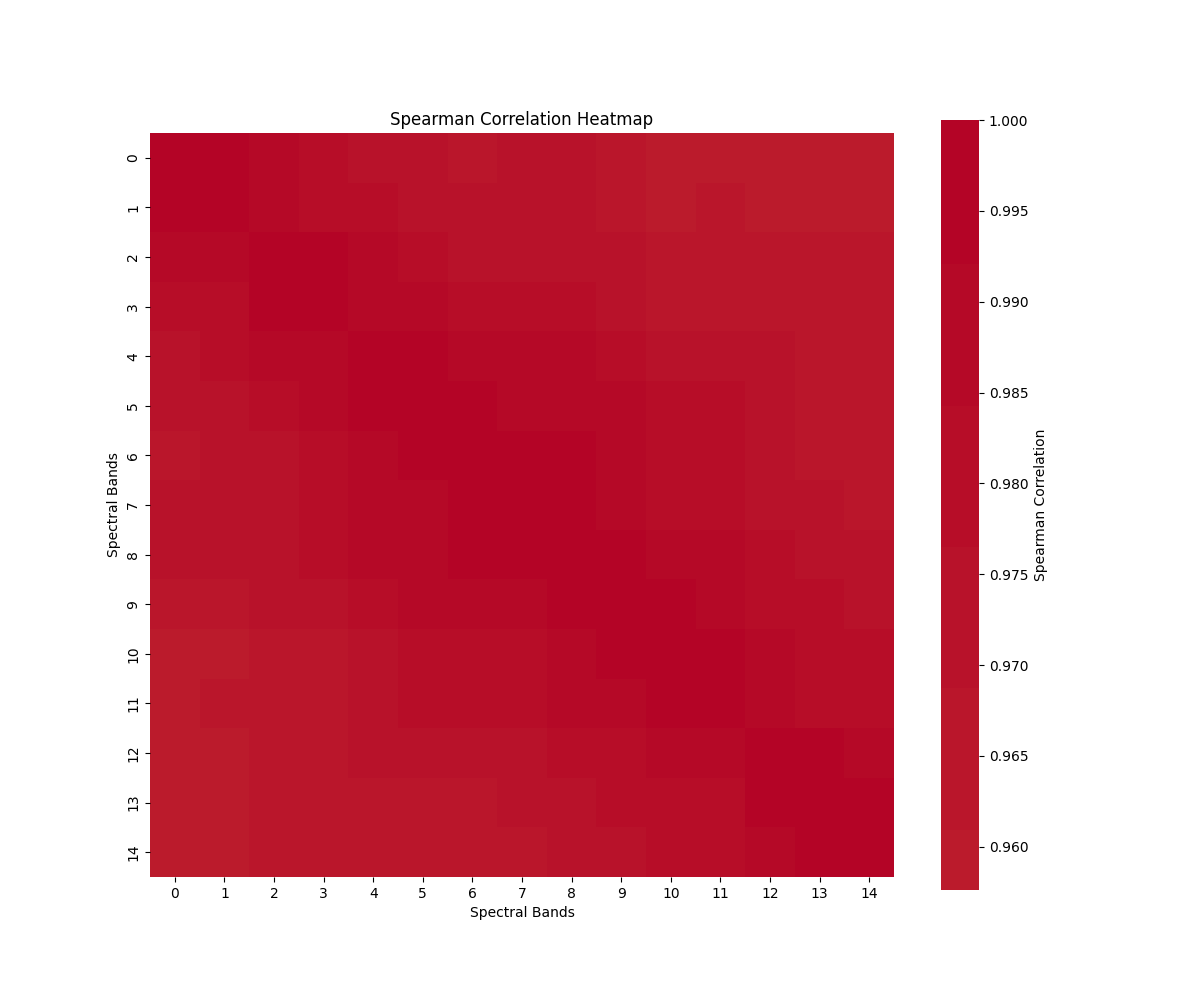}
        \caption{2D-CNN}
    \end{subfigure}
    \hfill
    \begin{subfigure}[b]{0.32\linewidth}
        \includegraphics[width=\linewidth]{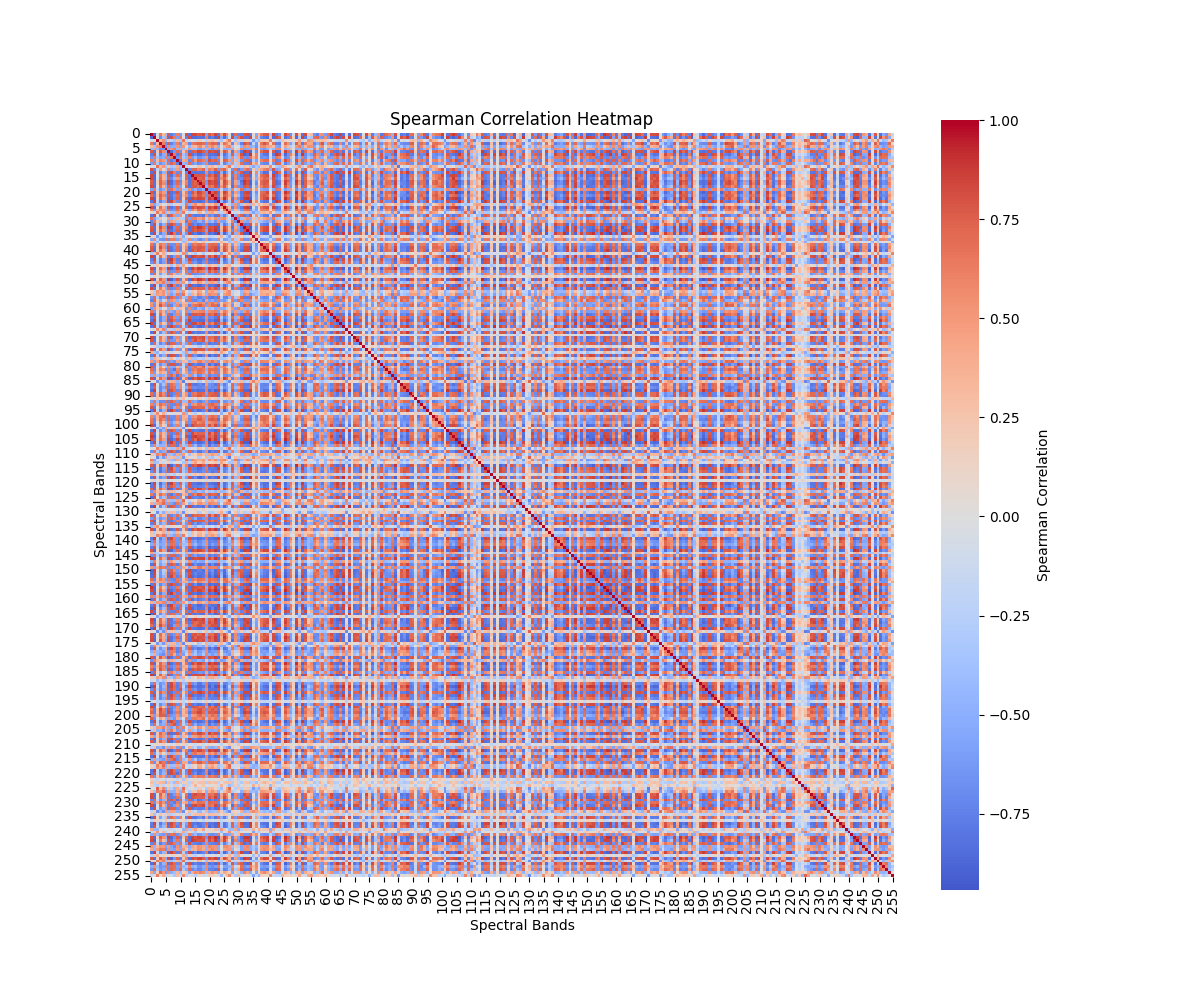}
        \caption{HiT}
    \end{subfigure}
    \hfill
    \begin{subfigure}[b]{0.32\linewidth}
        \includegraphics[width=\linewidth]{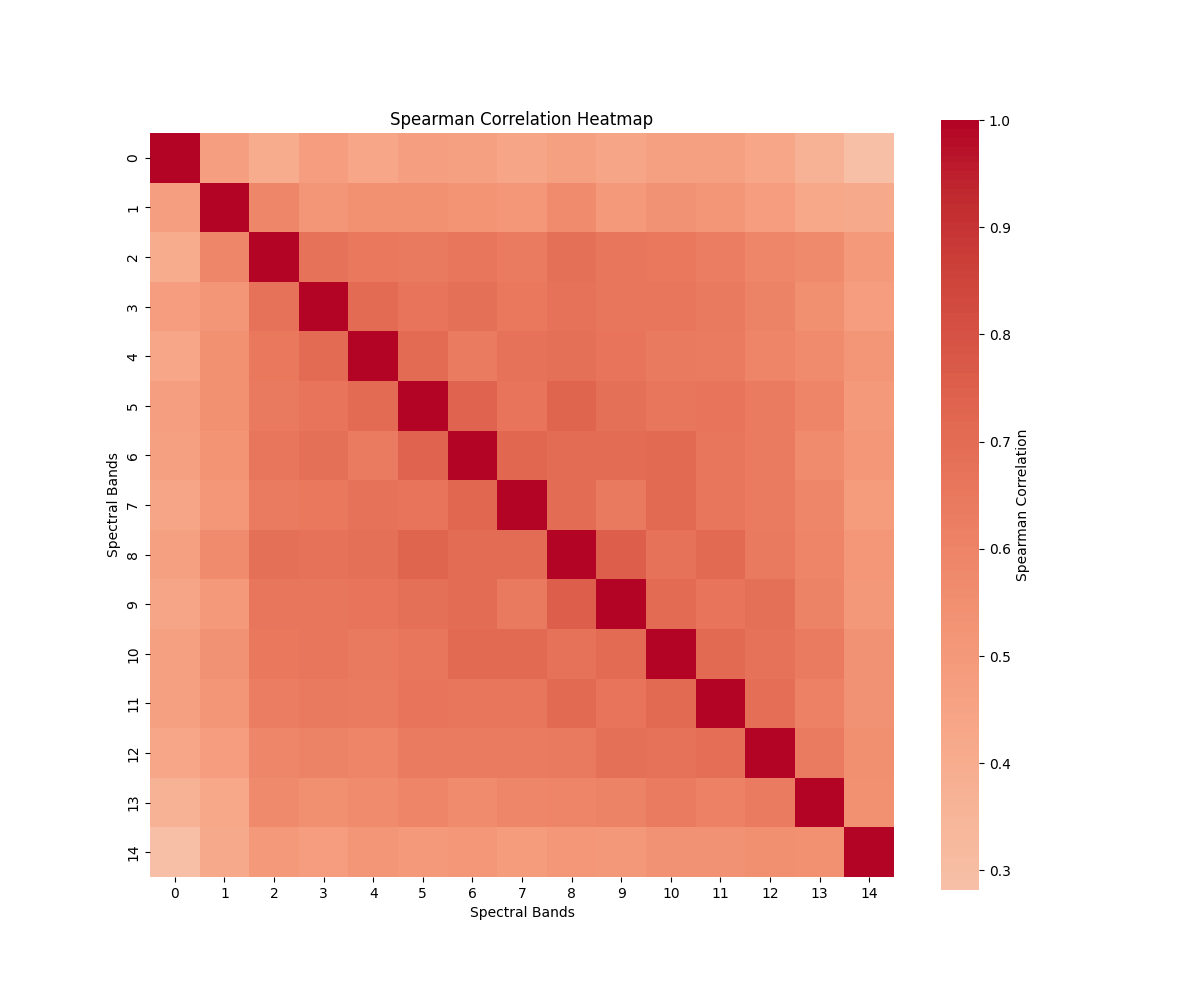}
        \caption{DCT-Mamba3D}
    \end{subfigure}
    \caption{Spearman correlation heatmaps on the Indian Pines dataset, comparing (a) 2D-CNN, (b) HiT (Transformer-based), and (c) DCT-Mamba3D.}
    \label{fig:heatmaps}
\end{figure}

Figure~\ref{fig:heatmaps} shows Spearman correlation heatmaps on the Indian Pines dataset for (a) 2D-CNN, (b) HiT, and (c) DCT-Mamba3D. In each heatmap, the x- and y-axes represent spectral bands, with high off-diagonal values indicating spectral redundancy. The 2D-CNN retains substantial redundancy, HiT reduces some but maintains redundancy in adjacent bands, while DCT-Mamba3D achieves marked decorrelation, contributing to improved classification performance.

\subsection{Classification Accuracy Comparison}
We compare the classification performance of DCT-Mamba3D with leading models, including 2D-CNN, 3D-CNN, HybridSN, ViT, HiT, MorphF, SSFTT, and MiM. As shown in Table~\ref{tab:indian_pines}, DCT-Mamba3D demonstrates effective performance in hyperspectral image classification, particularly in challenging cases.

Our method demonstrates particular advantages in challenging cases:
- \textit{Same Object, Different Spectra}: For classes like Corn-notill and Corn-mintill, DCT-Mamba3D substantially improves classification accuracy, effectively capturing intra-class spectral variability. The decorrelation provided by the 3D Spectral Decorrelation Module enables the model to distinguish subtle spectral variations within similar classes, resulting in higher F1-scores for these classes.
- \textit{Different Objects, Same Spectra}: For classes such as Buildings-Grass-Trees-Drives and Stone-Steel-Towers, DCT-Mamba3D outperforms other models, highlighting its ability to reduce spectral redundancy and improve separability in classes with similar spectral characteristics.

\subsubsection{t-SNE Visualization Analysis}

\begin{figure*}[htbp]
    \centering
    \begin{subfigure}[b]{0.32\textwidth}
        \includegraphics[width=\textwidth]{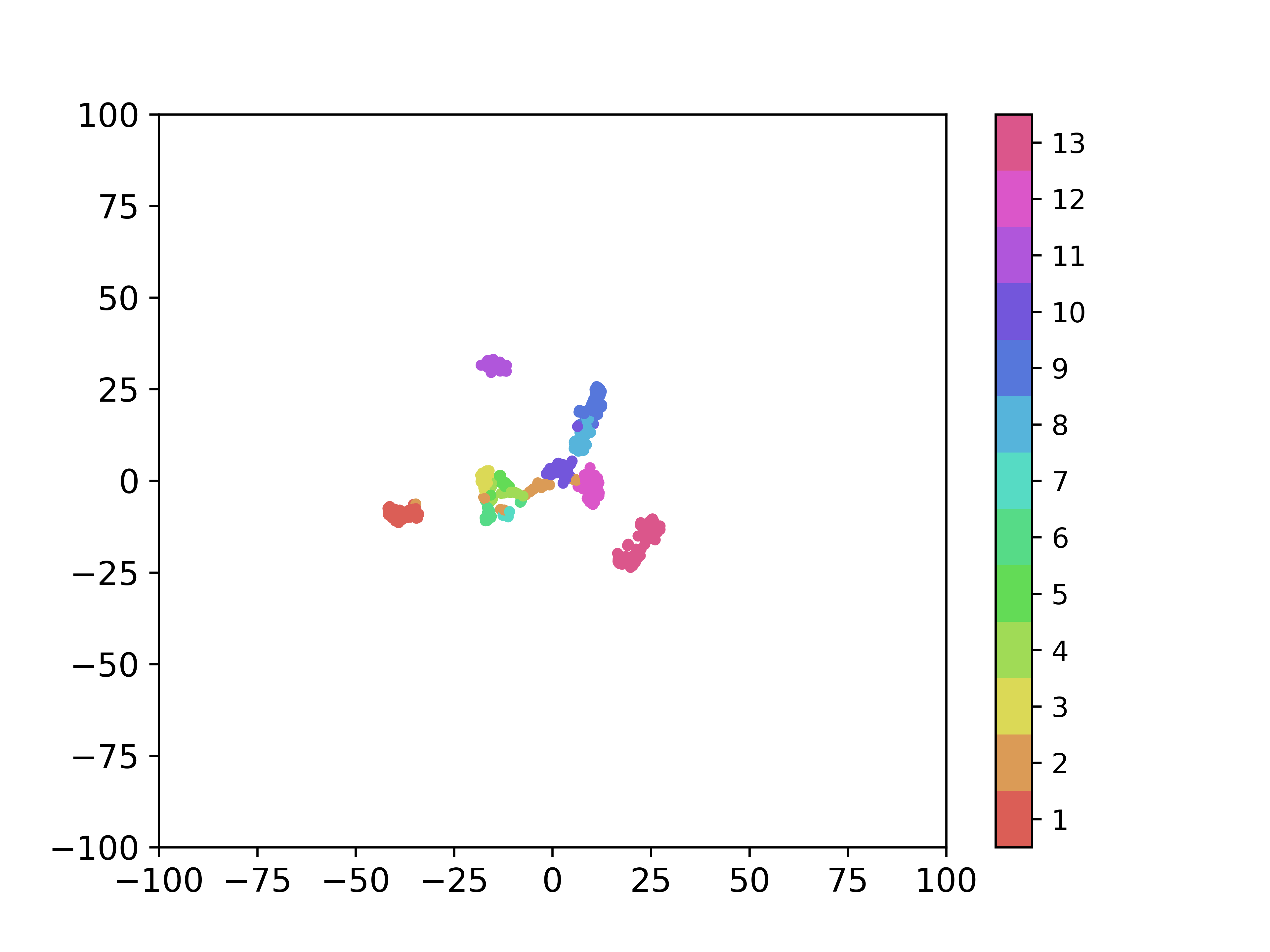}
        \caption{2D-CNN}
    \end{subfigure}
    \hfill
    \begin{subfigure}[b]{0.32\textwidth}
        \includegraphics[width=\textwidth]{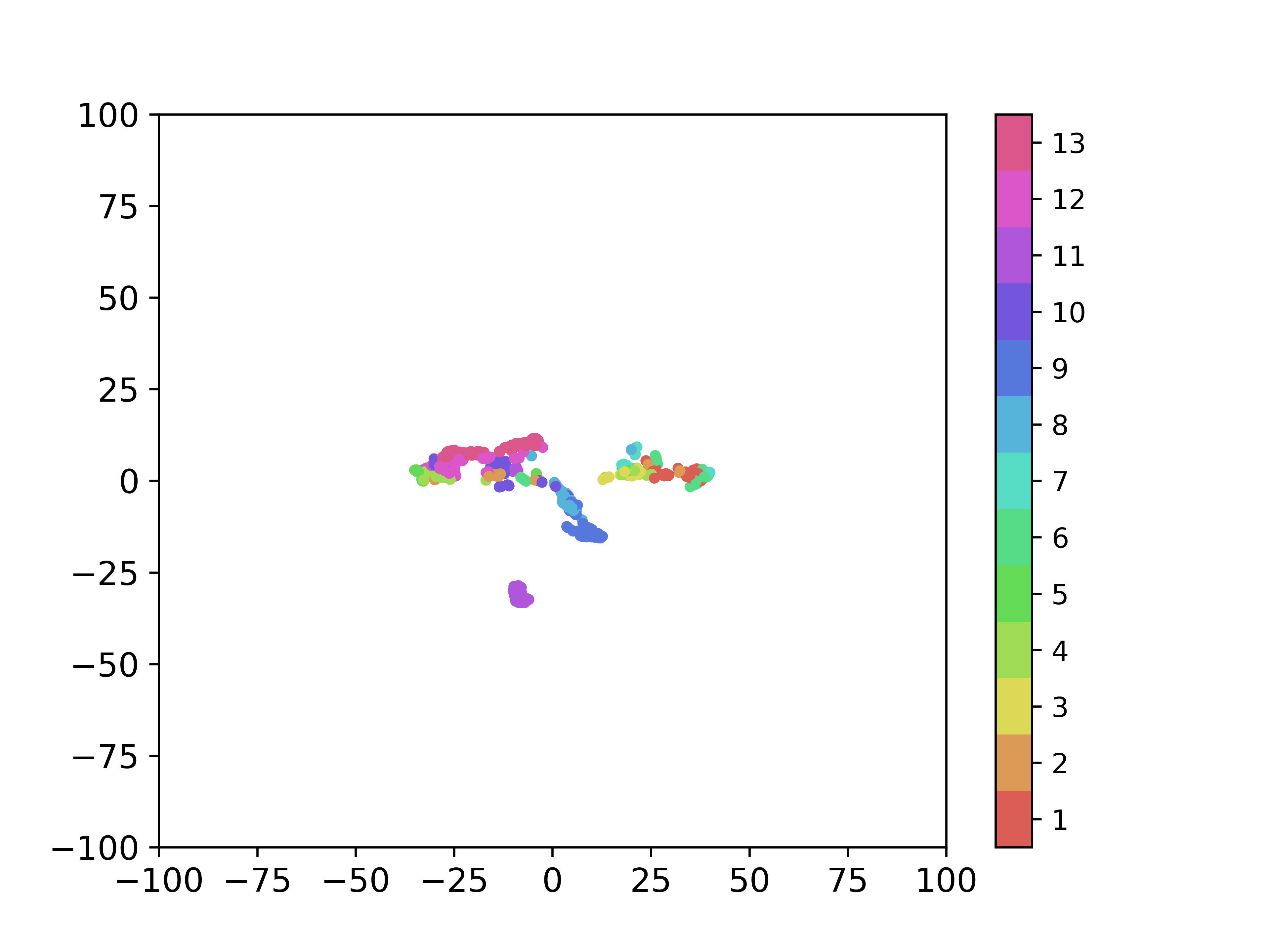}
        \caption{3D-CNN}
    \end{subfigure}
    \hfill
    \begin{subfigure}[b]{0.32\textwidth}
        \includegraphics[width=\textwidth]{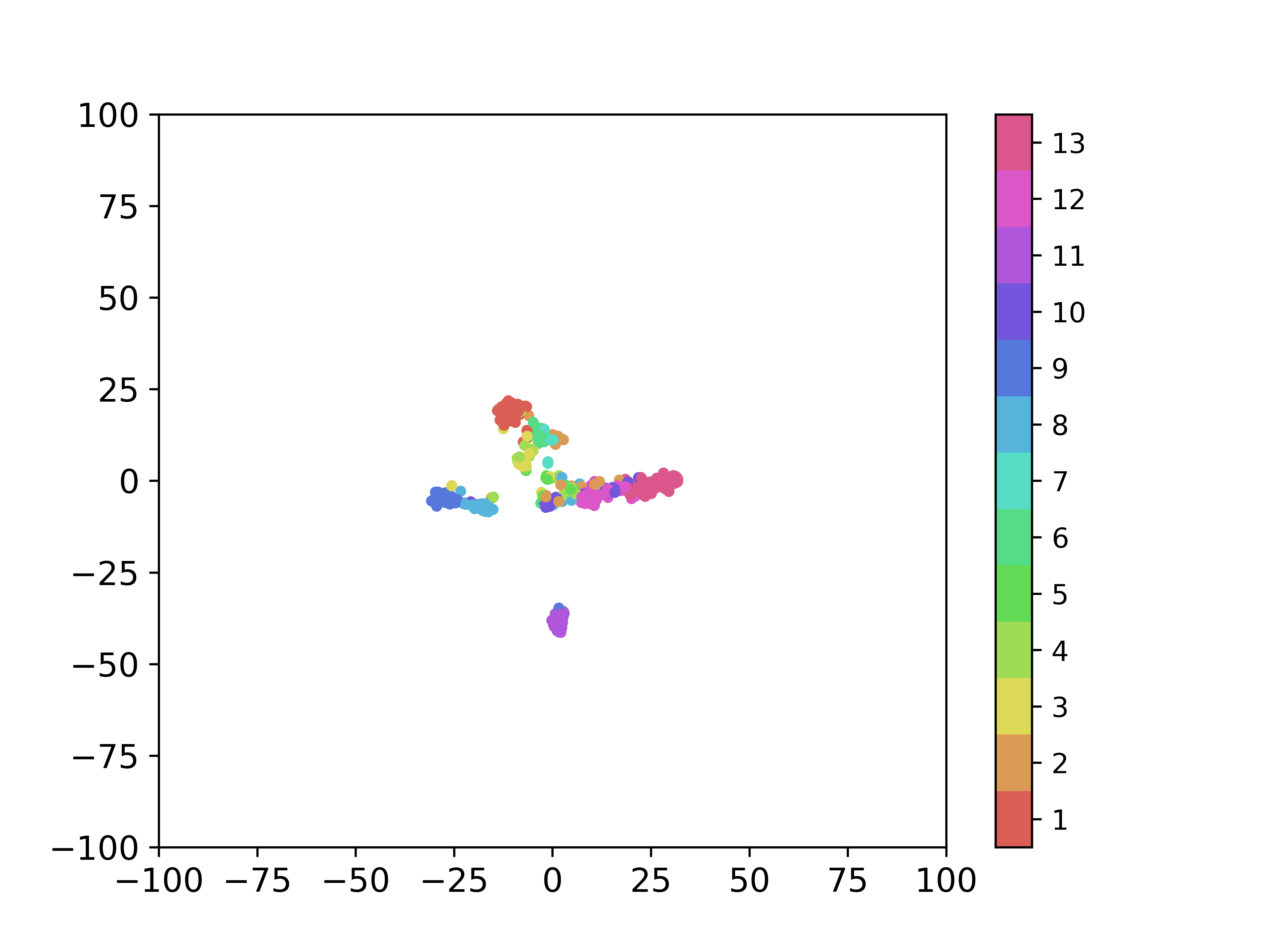}
        \caption{HiT}
    \end{subfigure}
    \\
    \begin{subfigure}[b]{0.32\textwidth}
        \includegraphics[width=\textwidth]{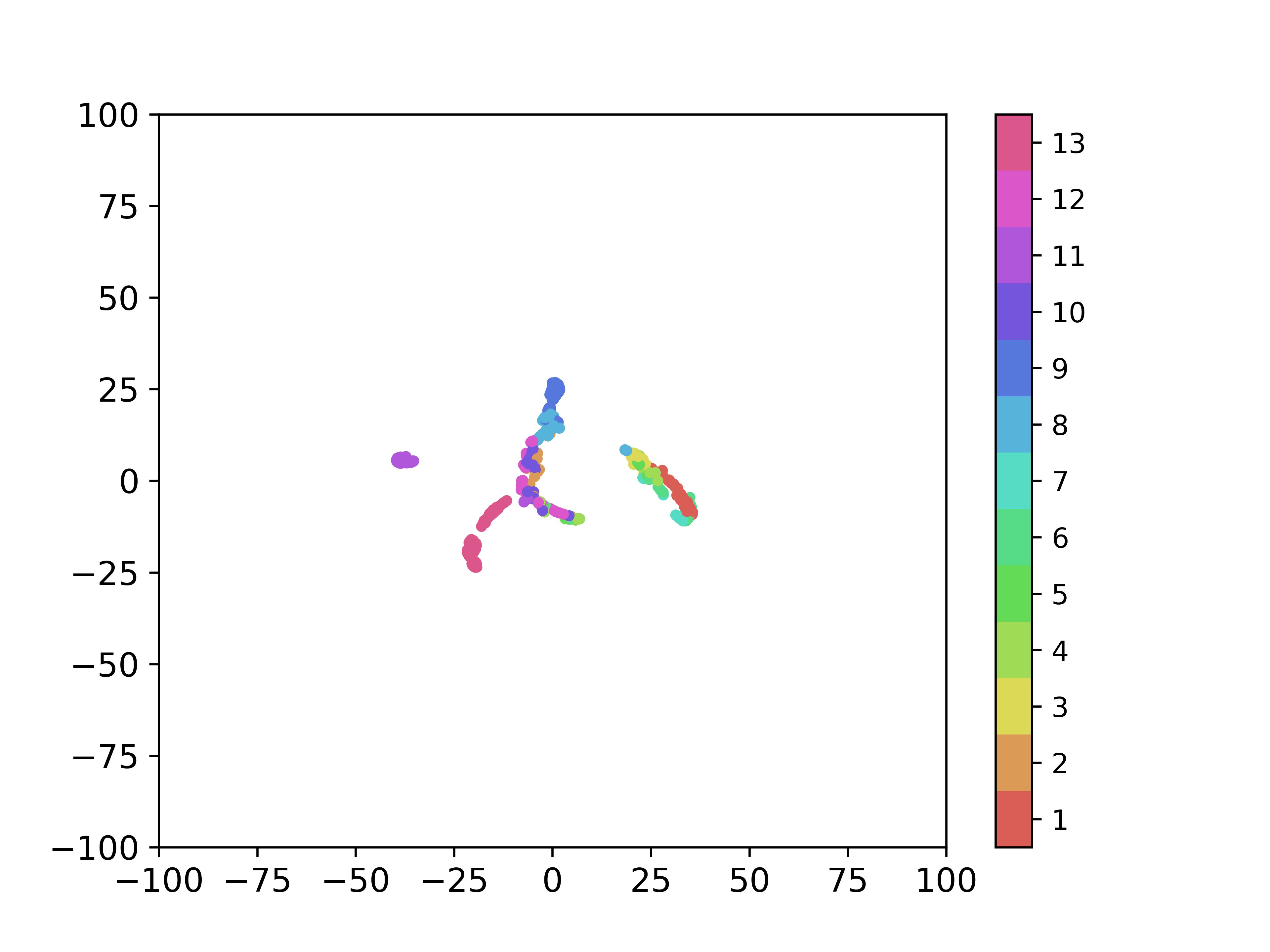}
        \caption{MorphF}
    \end{subfigure}
    \hfill
    \begin{subfigure}[b]{0.32\textwidth}
        \includegraphics[width=\textwidth]{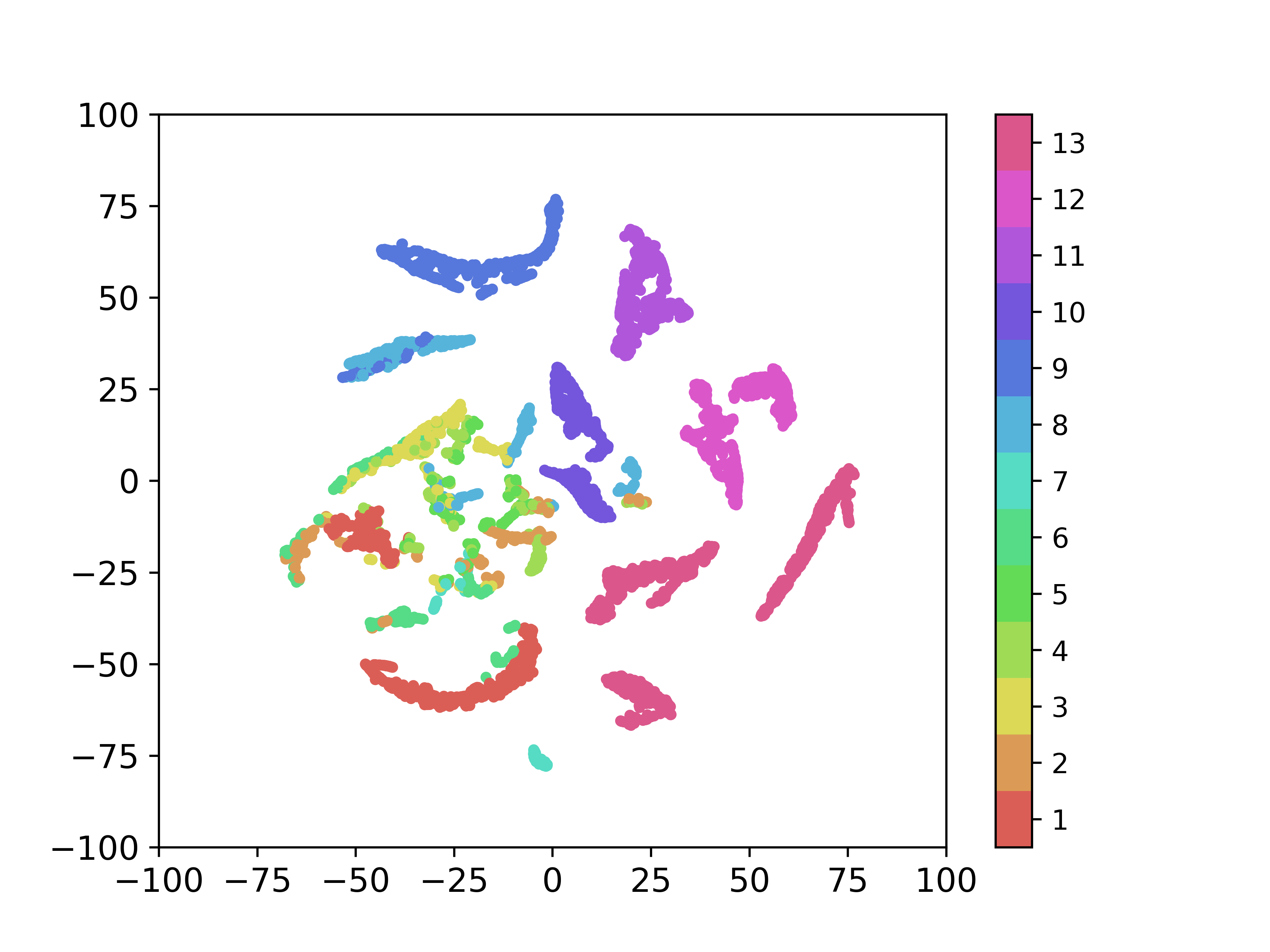}
        \caption{MiM}
    \end{subfigure}
    \hfill
    \begin{subfigure}[b]{0.32\textwidth}
        \includegraphics[width=\textwidth]{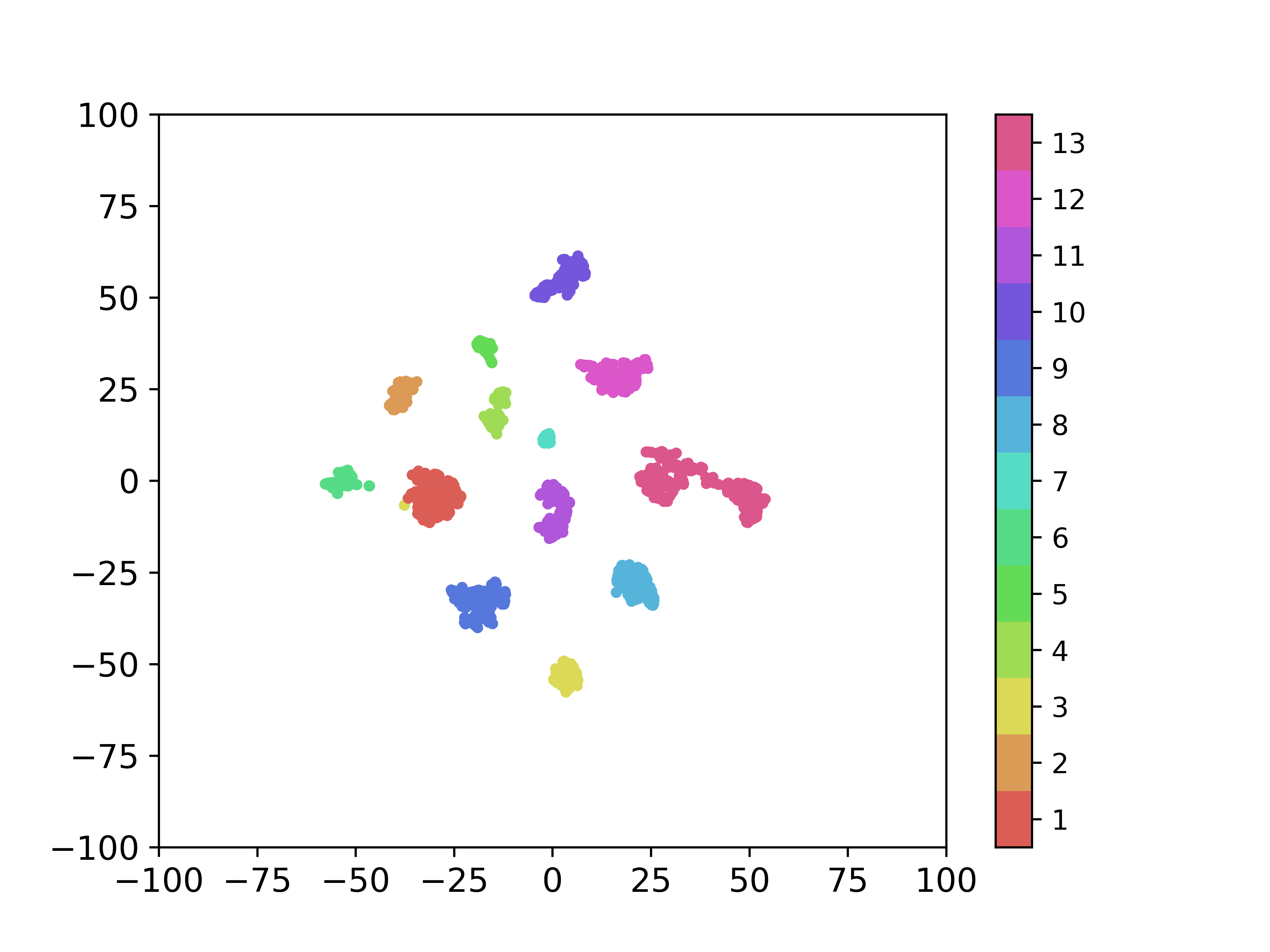}
        \caption{DCT-Mamba3D}
    \end{subfigure}
    \caption{t-SNE visualization of feature embeddings across models on the KSC dataset.}
    \label{fig:tsne_comparison}
\end{figure*}

We conducted a t-SNE visualization analysis to evaluate the effectiveness of DCT-Mamba3D in feature discrimination, as shown in Fig.~\ref{fig:tsne_comparison}. The results indicate that DCT-Mamba3D generates compact, well-separated clusters, demonstrating its robust capability in decorrelating complex spectral and spatial features. This enhanced decorrelation improves class separability by reducing interclass misclassification and strengthens in-class cohesion, addressing key challenges in hyperspectral imaging, such as “same object, different spectra” and “different objects, same spectra.” By leveraging 3D-SSDM, DCT-Mamba3D effectively minimizes spectral and spatial redundancies. It results in clear and distinct feature representations, particularly beneficial in scenarios with overlapping or redundant spectral information.

\begin{table*}[htbp]
\centering
\caption{Classification accuracy comparison on the Indian Pines dataset with 10\% training samples.}
\label{tab:indian_pines}
\scriptsize
\begin{tabular}{lccccccccc}
\toprule
\textbf{Class} & \textbf{2D-CNN} & \textbf{3D-CNN} & \textbf{HybridSN} & \textbf{ViT} & \textbf{HiT} & \textbf{MorphF} & \textbf{SSFTT} & \textbf{MiM} & \textbf{DCT-Mamba3D} \\
\midrule
Alfalfa & 92.82 & 70.82 & 34.67 & 76.55 & 0.00 & 79.58 & 89.95 & 76.20 & \textbf{95.19} \\
Corn-notill & 93.81 & 89.33 & 88.50 & 94.10 & 91.21 & 93.08 & 92.40 & 92.49 & \textbf{95.15} \\
Corn-mintill & 92.19 & 87.44 & 81.40 & 93.16 & 89.69 & 90.15 & 88.62 & 89.03 & \textbf{93.47} \\
Corn & 97.94 & 94.78 & 83.47 & \textbf{99.58} & 84.51 & 92.96 & 96.23 & 93.06 & 99.01 \\
Grass-pasture & 93.09 & 92.88 & 84.74 & 91.34 & 45.75 & \textbf{95.13} & 93.71 & 93.20 & 94.17 \\
Grass-trees & 95.65 & 94.41 & 82.42 & 89.85 & 94.06 & 95.80 & 94.43 & \textbf{96.01} & 94.72 \\
Grass-pasture-mowed & 7.94 & 0.00 & 1.21 & 0.00 & 0.00 & \textbf{66.25} & 56.16 & 32.43 & 16.11 \\
Hay-windrowed & 99.69 & 99.09 & 92.37 & 99.83 & \textbf{100.00} & 99.81 & 99.23 & 99.92 & 99.41 \\
Oats & \textbf{73.30} & 0.00 & 0.00 & 0.00 & 0.00 & 9.19 & 38.88 & 53.87 & 44.60 \\
Soybean-notill & 87.78 & 83.76 & 82.54 & \textbf{89.52} & 81.26 & 89.35 & 87.84 & 86.27 & 80.89 \\
Soybean-mintill & 96.26 & 94.33 & 93.02 & 96.58 & \textbf{96.92} & 96.71 & 96.08 & 96.27 & 96.77 \\
Soybean-clean & 91.80 & 89.17 & 81.26 & 91.69 & 87.08 & 88.93 & 87.39 & 84.99 & \textbf{93.85} \\
Wheat & \textbf{98.12} & 86.12 & 47.40 & 97.28 & 92.43 & 92.52 & 93.27 & 91.09 & 97.83 \\
Woods & 98.28 & 97.96 & 97.28 & 98.20 & 99.74 & 98.87 & \textbf{98.88} & 98.04 & 98.51 \\
Buildings-Grass-Trees-Drives & 97.82 & 92.51 & 74.80 & 98.05 & 87.90 & 95.08 & 94.88 & 92.95 & \textbf{98.74} \\
Stone-Steel-Towers & 52.74 & 51.18 & 15.74 & 34.79 & 0.00 & 55.80 & 34.81 & 15.01 & \textbf{68.36} \\
\midrule
\textbf{Accuracy (\%)} & 94.48 & 91.65 & 86.81 & 94.18 & 88.88 & 94.14 & 93.36 & 92.81 & \textbf{95.23} \\
\textbf{Kappa (\%)} & 93.69 & 90.45 & 84.87 & 93.35 & 87.24 & 93.30 & 92.42 & 91.78 & \textbf{94.55} \\
\bottomrule
\end{tabular}
\end{table*}

\begin{table*}[htbp]
\centering
\caption{Performance Comparison on Kennedy Space Center Dataset (10\% Training Samples)}
\label{tab:ksc}
\scriptsize % Adjust to make it fit without shrinking too much
\begin{tabular}{lrrrrrrrrr}
\toprule
\textbf{Class} & \textbf{2D-CNN} & \textbf{3D-CNN} & \textbf{HySN} & \textbf{ViT} & \textbf{HiT} & \textbf{MorphF} & \textbf{SSFTT} & \textbf{MiM} & \textbf{DCT-Mamba3D} \\
\midrule
Scrub                & 97.38 & 95.96 & 95.89 & 91.62 & 86.13 & 80.24 & 92.20 & 98.32 & \textbf{99.02} \\
Willow swamp         & 95.16 & 87.20 & 87.77 & 74.05 & \textbf{98.69} & 57.59 & 88.83 & 93.30 & 98.13 \\
Cabbage palm hammock & 97.78 & 91.63 & 94.18 & 85.03 & \textbf{99.17} & 20.20 & 49.84 & 93.98 & 97.99 \\
Cabbage palm/oak ham & 92.82 & 83.38 & 89.29 & 93.50 & 98.36 & 52.04 & 82.66 & 74.07 & \textbf{99.84} \\
Slash pine           & 89.94 & 78.78 & 78.63 & 94.89 & 98.83 & 70.67 & 95.07 & 21.99 & \textbf{99.94} \\
Oak/broadleaf ham    & 94.68 & 93.27 & 90.85 & 93.21 & \textbf{99.74} & 41.96 & 90.31 & 93.72 & 98.72 \\
Hardwood swamp       & 98.79 & 99.32 & 99.05 & 99.50 & 99.02 & 25.76 & 66.07 & 91.61 & \textbf{100.00} \\
Graminoid marsh      & 94.66 & 97.62 & 96.55 & 89.51 & 96.78 & 67.25 & 85.22 & \textbf{99.62} & 98.74 \\
Spartina marsh       & 97.49 & 99.70 & 99.40 & 95.99 & \textbf{99.99} & 73.41 & 81.59 & 99.89 & 99.94 \\
Cattail marsh        & 97.17 & 97.37 & 98.36 & 97.75 & 99.03 & 89.24 & \textbf{100.00} & \textbf{100.00} & 99.98 \\
Salt marsh           & 99.65 & \textbf{99.98} & 98.06 & \textbf{100.00} & \textbf{100.00} & 97.67 & 98.28 & \textbf{100.00} & \textbf{100.00} \\
Mud flats            & 98.28 & 97.85 & 99.08 & 99.65 & 99.17 & 95.03 & \textbf{100.00} & \textbf{100.00} & \textbf{100.00} \\
Water                & \textbf{100.00} & \textbf{100.00} & \textbf{100.00} & \textbf{100.00} & 99.86 & \textbf{100.00} & \textbf{100.00} & \textbf{100.00} & \textbf{100.00} \\
\midrule
\textbf{Accuracy (\%)} & 97.31 & 95.99 & 96.28 & 94.77 & 95.51 & 80.36 & 91.99 & 95.80 & \textbf{99.50} \\
\textbf{Kappa (\%)}    & 97.01 & 95.54 & 95.86 & 94.16 & 95.03 & 77.82 & 91.02 & 95.32 & \textbf{99.44} \\
\bottomrule
\end{tabular}
\end{table*}

\begin{table*}[htbp]
\centering
\caption{Comparison with Leading Transformer and Mamba Models on Houston2013 Dataset (10\% Training Samples)}
\label{tab:houston}
\scriptsize
\begin{tabular}{lccccccccc}
\toprule
\textbf{Class} & \textbf{2D-CNN} & \textbf{3D-CNN} & \textbf{HybridSN} & \textbf{ViT} & \textbf{HiT} & \textbf{MorphF} & \textbf{SSFTT} & \textbf{MiM} & \textbf{DCT-Mamba3D} \\
\midrule
Unclassified & 95.45 & 93.10 & 92.46 & 93.45 & 94.63 & 93.78 & 90.18 & 93.36 & \textbf{96.60} \\
Healthy Grass & 96.45 & 90.56 & 88.22 & 87.16 & 90.15 & 94.20 & 92.70 & 95.09 & \textbf{97.78} \\
Stressed Grass & 99.32 & 97.35 & 97.84 & 97.29 & 98.93 & 99.18 & \textbf{99.52} & 99.03 & 98.89 \\
Synthetic Grass & 92.42 & 89.78 & 81.79 & 85.45 & 87.57 & \textbf{97.76} & 93.53 & 91.13 & 96.54 \\
Soil & 99.57 & 97.53 & 97.47 & 99.56 & 99.66 & 99.74 & 99.33 & \textbf{99.90} & 99.30 \\
Water & 95.02 & 86.73 & 93.09 & 90.34 & 88.05 & 94.86 & \textbf{93.44} & 91.60 & 93.01 \\
Residential & 93.96 & 86.51 & 86.07 & 89.16 & 90.17 & 97.10 & 95.40 & 90.88 & \textbf{98.10} \\
Commercial & 96.59 & 89.12 & 92.09 & 96.86 & 96.95 & 98.21 & 96.59 & \textbf{99.39} & 97.38 \\
Road & 94.59 & 84.52 & 76.40 & 90.09 & 89.77 & 96.95 & 95.06 & 93.98 & \textbf{97.42} \\
Highway & 97.30 & 94.23 & 94.67 & 99.17 & 97.88 & 99.20 & 99.22 & 99.73 & \textbf{99.90} \\
Railway & 99.41 & 87.33 & 82.52 & 98.29 & 99.50 & 99.85 & 99.56 & 98.10 & \textbf{99.96} \\
Parking Lot 1 & 97.78 & 95.51 & 96.37 & 98.04 & 98.54 & 99.45 & 98.21 & \textbf{99.24} & 99.17 \\
Parking Lot 2 & 96.27 & 88.31 & 87.86 & 94.16 & 94.78 & \textbf{97.38} & 97.69 & 96.77 & 96.94 \\
Tennis Court & 99.92 & 98.84 & 96.61 & 99.85 & \textbf{99.95} & 99.87 & 99.51 & 99.78 & 99.85 \\
Running Track & 98.92 & 95.49 & 94.23 & 96.77 & 98.75 & 98.96 & \textbf{99.15} & 98.81 & 98.58 \\
\midrule
\textbf{Accuracy (\%)} & 96.66 & 91.36 & 90.09 & 94.09 & 94.87 & 97.75 & 96.37 & 96.32 & \textbf{98.15} \\
\textbf{Kappa (\%)} & 96.39 & 90.65 & 89.29 & 93.61 & 94.45 & 97.57 & 96.08 & 96.02 & \textbf{98.00} \\
\bottomrule
\end{tabular}
\end{table*}

\subsection{Training Loss Comparison}

\begin{figure}[htbp]
    \centering
        \includegraphics[width=\linewidth]{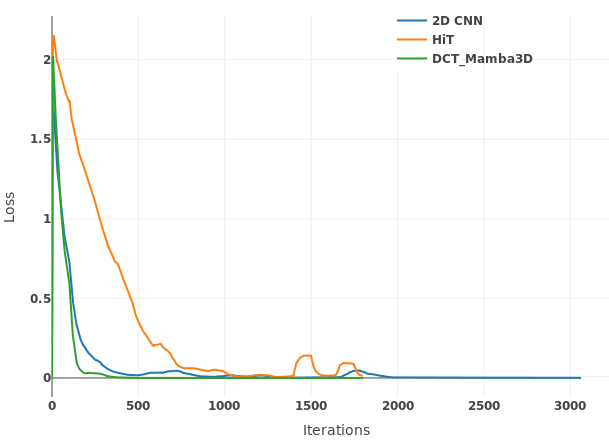}
    \caption{Training loss comparison between 2D-CNN, HiT, and the DCT-Mamba3D model. }
    \label{training_loss}
\end{figure}

To further demonstrate the efficiency of our proposed model, we compare the training loss curves of the DCT-Mamba3D model with those of a 2D-CNN and HiT, as shown in Figure~\ref{training_loss}. Each model's training loss was recorded across the training iterations to analyze convergence behavior. The results clearly show that the DCT-Mamba3D model achieves a faster and more stable convergence compared to the 2D-CNN and HiT baselines.

\begin{table}[H]
\centering
\caption{Ablation Study Results on the Indian Pines Dataset, showing Overall Accuracy (OA), Average Accuracy (AA), and Kappa Score for each configuration.}
\label{tab:ablation_summary}
\resizebox{\linewidth}{!}{
\begin{tabular}{|l|c|c|c|}
\hline
\textbf{Configuration} & OA (\%) & AA (\%) & Kappa (\%) \\
\hline
3D-SDM Only & 93.55 & 82.63 & 92.65 \\
3D-MambaNet Only & 94.97 & 85.56 & 94.26 \\
No GRE & 94.62 & 85.08 & 93.87 \\
Full DCT-Mamba3D & 95.23 & 86.81 & 94.55 \\
\hline
\end{tabular}
}
\end{table}

\begin{table}[H]
\centering
\caption{DCT-Mamba3D Performance on Different Datasets and Sample Sizes, showing Overall Accuracy (OA) and Kappa Score for each configuration.}
\label{tab:data_quantity}
\resizebox{\linewidth}{!}{
\begin{tabular}{|c|c|c|c|}
\hline
\textbf{Dataset} & \textbf{Sample Size (\%)} & \textbf{OA (\%)} & \textbf{Kappa (\%)} \\
\hline
Indian Pines & 1\% & 49.58 & 40.68 \\
Indian Pines & 3\% & 77.81 & 74.59 \\
Indian Pines & 5\% & 82.79 & 80.21 \\
Indian Pines & 7\% & 86.67 & 84.69 \\
Indian Pines & 10\% & 95.23 & 94.55 \\
KSC & 1\% & 61.64 & 56.22 \\
KSC & 3\% & 78.18 & 75.45 \\
KSC & 5\% & 89.21 & 87.95 \\
KSC & 7\% & 93.54 & 92.80 \\
KSC & 10\% & 99.50 & 99.44 \\
Houston2013 & 1\% & 68.75 & 66.17 \\
Houston2013 & 3\% & 86.20 & 85.08 \\
Houston2013 & 5\% & 92.09 & 91.45 \\
Houston2013 & 7\% & 94.46 & 94.01 \\
Houston2013 & 10\% & 98.15 & 98.00 \\
\hline
\end{tabular}
}
\end{table}

\begin{table}[H]
\centering
\caption{Complexity Analysis on Indian Pines Dataset}
\label{complexity}
\begin{tabular}{|c|c|c|}
\hline
\textbf{Method} & FLOPS & Parameters (M) \\
\hline
3D-CNN & 0.50 & 0.50 \\
HybridSN & 5.3 & 4.32 \\
ViT & 0.68 & 13.2 \\
HiT & 11.93 & 20.94 \\
MorphF & 0.17 & 0.24 \\
SSFTT & 0.24 & 0.93 \\
MiM & 0.50 & 0.18 \\
DCT-Mamba3D & 4.02 & 19.56 \\
\hline
\end{tabular}
\end{table}

The training loss curves highlight a significant advantage of the DCT-Mamba3D model in terms of convergence speed. It achieves near-zero training loss in fewer iterations than both 2D-CNN and HiT, showcasing its efficient feature extraction and decorrelation capabilities in the frequency domain. This rapid convergence not only reduces training time but also demonstrates the robustness of the model in handling complex spatial-spectral dependencies, further supporting its effectiveness in hyperspectral image classification.

\subsection{Ablation Study and Complexity Analysis}

We conducted an ablation study to evaluate the contributions of key components in the DCT-Mamba3D model. Three configurations were tested:

\begin{enumerate}
    \item \textit{3D-SDM Only}: Excludes the 3D-MambaNet, focusing on spectral decorrelation, achieving OA of 93.55\%, AA of 82.63\%, and Kappa score of 92.65\%.
    \item \textit{3D-MambaNet Only}: Excludes the 3D-SDM, focusing on spatial feature extraction, achieving OA of 94.97\%, AA of 85.56\%, and Kappa score of 94.26\%.
    \item \textit{No GRE (Global Residual Enhancement)}: Excludes the GRE, achieving OA of 94.62\%, AA of 85.08\%, and Kappa score of 93.87\%.
\end{enumerate}

Table~\ref{tab:ablation_summary} summarizes the OA, AA, and Kappa scores for each configuration, alongside the full DCT-Mamba3D model. The results highlight the model's integrated spatial-spectral feature extraction advantage, particularly for "same object, different spectra" and "different objects, same spectra" cases.

To further examine DCT-Mamba3D’s performance, we evaluated it on the Indian Pines, KSC, and Houston2013 datasets with varying training sample sizes, particularly under limited training samples, as shown in Table~\ref{tab:data_quantity}. The results indicate that DCT-Mamba3D maintains high OA and Kappa scores even with limited samples (e.g., 1\% and 3\%), underscoring its robust feature extraction capabilities. This effectiveness can be attributed to the 3D-SSDM’s spectral-spatial decorrelation, which reduces redundancy and enhances feature separability. As sample size increases, the model consistently improves in both OA and Kappa, demonstrating its scalability and robustness across varying data conditions, making it particularly suitable for challenging HSI classification tasks.

\subsubsection{Complexity Analysis}

We evaluated the computational complexity of DCT-Mamba3D relative to baseline models on the Indian Pines dataset, as detailed in Table~\ref{complexity}, focusing on FLOPS and parameter counts. While DCT-Mamba3D incurs a higher computational cost than CNN models, this is offset by substantial performance gains. Compared to other Mamba-based architectures, such as MiM, DCT-Mamba3D demonstrates an optimized balance of complexity, accuracy, and decorrelation effectiveness, making it well-suited for applications requiring both precision and computational efficiency.

\section{Discussion}
\label{sec:discussion}

Our \textbf{DCT-Mamba3D} model addresses HSI classification challenges by leveraging spectral-spatial decorrelation and effective feature extraction through its integrated modules: \textbf{3D-SSDM}, \textbf{3D-Mamba}, and \textbf{GRE}.

\textbf{3D-SSDM for Spectral Decorrelation and Feature Extraction:}  
Using 3D DCT basis functions, 3D-SSDM reduces spectral and spatial redundancy while extracting critical features across dimensions. Its enhanced decorrelation improves the model’s ability to manage cases of high spectral similarity, such as in 'different objects, same spectra' scenarios. Ablation studies show that the 3D-SSDM alone achieves competitive accuracy, underscoring its effectiveness in isolating relevant features.

\textbf{3D-Mamba for Enhanced Spatial-Spectral Dependencies:}  
3D-Mamba effectively captures spatial-spectral dependencies through bidirectional state-space layers, particularly aiding in differentiating subtle spectral variations, as in "same object, different spectra" scenarios. This structure balances complexity while maintaining efficient feature interaction across spatial and spectral dimensions.

\textbf{GRE for Stability and Robust Training:}  
GRE integrates residual spatial-spectral features, stabilizing feature representation across layers. This stability accelerates convergence, contributing to DCT-Mamba3D’s efficient training times and robustness during learning, despite its complex architecture.

\textbf{Performance with Limited Data and Complexity Efficiency:}  
DCT-Mamba3D demonstrates strong feature discrimination even with limited training data, achieving high accuracy and Kappa scores at smaller sample sizes. Its balanced computational complexity, as shown in Table~\ref{complexity}, allows for superior accuracy compared to other models, with manageable FLOPS and parameter counts, making it suitable for resource-constrained applications.

\section{Conclusion}
\label{sec:conclusion}

In this paper, we proposed \textbf{DCT-Mamba3D}, a framework for hyperspectral image (HSI) classification designed to address spectral redundancy and complex spatial-spectral dependencies. The architecture integrates three core modules: a \textbf{3D Spatial-Spectral Decorrelation Module (3D-SSDM)} that utilizes a 3D DCT basis functions to reduce redundancy and enhance feature clarity; a \textbf{3D-Mamba module} that captures intricate spatial-spectral relationships through a bidirectional state-space model; and a \textbf{Global Residual Enhancement (GRE) module} to stabilize feature representation for improved robustness and convergence. Our experiments on benchmark HSI datasets demonstrated that DCT-Mamba3D surpasses state-of-the-art methods, particularly in challenging scenarios with high spectral similarity or spectral variability.

\textbf{Future Work:}  
Future improvements could focus on exploring adaptive frequency selection to further enhance the generalizability of DCT-Mamba3D across diverse HSI datasets, potentially making it more robust to varying data characteristics and environmental conditions.

% References
{
    \small
    \bibliographystyle{ieeenat_fullname}
    \bibliography{main}

\begin{thebibliography}{36}
\providecommand{\natexlab}[1]{#1}
\providecommand{\url}[1]{\texttt{#1}}
\expandafter\ifx\csname urlstyle\endcsname\relax
  \providecommand{\doi}[1]{doi: #1}\else
  \providecommand{\doi}{doi: \begingroup \urlstyle{rm}\Url}\fi

\bibitem[Ahmad et~al.(2024)Ahmad, Usama, and Mazzara]{ahmad2024wavemamba}
Muhammad Ahmad, Muhammad Usama, and Manual Mazzara.
\newblock Wavemamba: Spatial-spectral wavelet mamba for hyperspectral image
  classification.
\newblock \emph{arXiv preprint arXiv:2408.01231}, 2024.

\bibitem[Ahmed et~al.(1974)Ahmed, Natarajan, and Rao]{ahmed1974discrete}
Nasir Ahmed, T\_ Natarajan, and Kamisetty~R Rao.
\newblock Discrete cosine transform.
\newblock \emph{IEEE Transactions on Computers}, 100\penalty0 (1):\penalty0
  90--93, 1974.

\bibitem[Bendoumi et~al.(2014)Bendoumi, He, and Mei]{be2014tgrs}
Mohamed~Amine Bendoumi, Mingyi He, and Shaohui Mei.
\newblock Hyperspectral image resolution enhancement using high-resolution
  multispectral image based on spectral unmixing.
\newblock \emph{IEEE Transactions on Geoscience and Remote Sensing},
  52\penalty0 (10):\penalty0 6574--6583, 2014.

\bibitem[Chen et~al.(2016)Chen, Jiang, Li, Jia, and Ghamisi]{chen2016deep}
Yushi Chen, Hanlu Jiang, Chunyang Li, Xiuping Jia, and Pedram Ghamisi.
\newblock Deep feature extraction and classification of hyperspectral images
  based on convolutional neural networks.
\newblock \emph{IEEE Transactions on Geoscience and Remote Sensing},
  54\penalty0 (10):\penalty0 6232--6251, 2016.

\bibitem[Deng et~al.(2023)Deng, Deng, Wu, Ran, Hong, and Vivone]{deng2023psrt}
Shang-Qi Deng, Liang-Jian Deng, Xiao Wu, Ran Ran, Danfeng Hong, and Gemine
  Vivone.
\newblock Psrt: Pyramid shuffle-and-reshuffle transformer for multispectral and
  hyperspectral image fusion.
\newblock \emph{IEEE Transactions on Geoscience and Remote Sensing},
  61:\penalty0 1--15, 2023.

\bibitem[Dosovitskiy et~al.(2020)Dosovitskiy, Beyer, Kolesnikov, Weissenborn,
  Zhai, Unterthiner, Dehghani, Minderer, Heigold, Gelly,
  et~al.]{dosovitskiy2020image}
Alexey Dosovitskiy, Lucas Beyer, Alexander Kolesnikov, Dirk Weissenborn,
  Xiaohua Zhai, Thomas Unterthiner, Mostafa Dehghani, Matthias Minderer, Georg
  Heigold, Sylvain Gelly, et~al.
\newblock An image is worth 16x16 words: Transformers for image recognition at
  scale.
\newblock \emph{arXiv preprint arXiv:2010.11929}, 2020.

\bibitem[Feng et~al.(2024)Feng, Wang, Zhang, Jia, and Yin]{feng2024cat}
Jiaqi Feng, Qixiong Wang, Guangyun Zhang, Xiuping Jia, and Jihao Yin.
\newblock Cat: Center attention transformer with stratified spatial-spectral
  token for hyperspectral image classification.
\newblock \emph{IEEE Transactions on Geoscience and Remote Sensing}, 2024.

\bibitem[He et~al.(2017)He, Li, Liu, and Li]{he2017recent}
Lin He, Jun Li, Chenying Liu, and Shutao Li.
\newblock Recent advances on spectral--spatial hyperspectral image
  classification: An overview and new guidelines.
\newblock \emph{IEEE Transactions on Geoscience and Remote Sensing},
  56\penalty0 (3):\penalty0 1579--1597, 2017.

\bibitem[Hong et~al.(2024)Hong, Zhang, Li, Li, Li, Yao, Yokoya, Li, Ghamisi,
  Jia, et~al.]{hong2024spectralgpt}
Danfeng Hong, Bing Zhang, Xuyang Li, Yuxuan Li, Chenyu Li, Jing Yao, Naoto
  Yokoya, Hao Li, Pedram Ghamisi, Xiuping Jia, et~al.
\newblock Spectralgpt: Spectral remote sensing foundation model.
\newblock \emph{IEEE Transactions on Pattern Analysis and Machine
  Intelligence}, 2024.

\bibitem[Jia et~al.(2022)Jia, Liao, Xu, Li, Zhu, Sun, Jia, and Li]{Jia2022tgrs}
Sen Jia, Jianhui Liao, Meng Xu, Yan Li, Jiasong Zhu, Weiwei Sun, Xiuping Jia,
  and Qingquan Li.
\newblock {3D} {Gabor} convolutional neural network for hyperspectral image
  classification.
\newblock \emph{IEEE Transactions on Geoscience and Remote Sensing},
  60:\penalty0 1--16, 2022.

\bibitem[Li et~al.(2012)Li, Bioucas-Dias, and Plaza]{li2013spectral}
Jun Li, Jos{\'e}~M Bioucas-Dias, and Antonio Plaza.
\newblock Spectral–spatial hyperspectral image segmentation using subspace
  multinomial logistic regression and markov random fields.
\newblock \emph{IEEE Transactions on Geoscience and Remote Sensing},
  50\penalty0 (3):\penalty0 809--823, 2012.

\bibitem[Li et~al.(2024{\natexlab{a}})Li, Fu, Zhang, Liu, Dou, Yan, and
  Zhang]{li2024latent}
Miaoyu Li, Ying Fu, Tao Zhang, Ji Liu, Dejing Dou, Chenggang Yan, and Yulun
  Zhang.
\newblock Latent diffusion enhanced rectangle transformer for hyperspectral
  image restoration.
\newblock \emph{IEEE Transactions on Pattern Analysis and Machine
  Intelligence}, 2024{\natexlab{a}}.

\bibitem[Li et~al.(2024{\natexlab{b}})Li, Luo, Zhang, Wang, and
  Du]{li2024mambahsi}
Yapeng Li, Yong Luo, Lefei Zhang, Zengmao Wang, and Bo Du.
\newblock Mambahsi: Spatial-spectral mamba for hyperspectral image
  classification.
\newblock \emph{IEEE Transactions on Geoscience and Remote Sensing},
  2024{\natexlab{b}}.

\bibitem[Paoletti et~al.(2019)Paoletti, Haut, Plaza, and
  Plaza]{paoletti2019deep}
Matteo~E Paoletti, Jean~M Haut, Javier Plaza, and Antonio Plaza.
\newblock Deep learning classifiers for hyperspectral imaging: A review.
\newblock \emph{ISPRS Journal of Photogrammetry and Remote Sensing},
  158:\penalty0 279--317, 2019.

\bibitem[Qiao and Huang(2023)]{qiao2023dual}
Xin Qiao and Weimin Huang.
\newblock A dual frequency transformer network for hyperspectral image
  classification.
\newblock \emph{IEEE Journal of Selected Topics in Applied Earth Observations
  and Remote Sensing}, pages 1--12, 2023.

\bibitem[Roy et~al.(2020)Roy, Krishna, and Dubey]{roy2019hybridsn}
Swapnil~Karun Roy, Gopal Krishna, and Shiv~Ram Dubey.
\newblock {HybridSN: Exploring 3D–2D CNN Feature Hierarchy for Hyperspectral
  Image Classification}.
\newblock \emph{IEEE Geoscience and Remote Sensing Letters}, 17\penalty0
  (2):\penalty0 277--281, 2020.

\bibitem[Roy et~al.(2023)Roy, Deria, Shah, Haut, Du, and
  Plaza]{roy2023spectral}
Soumya~Kanti Roy, Akash Deria, Chirag Shah, Jean-Marc Haut, Qian Du, and
  Antonio Plaza.
\newblock Spectral--spatial morphological attention transformer for
  hyperspectral image classification.
\newblock \emph{IEEE Transactions on Geoscience and Remote Sensing},
  61:\penalty0 1--15, 2023.

\bibitem[Scheibenreif et~al.(2023)Scheibenreif, Mommert, and
  Borth]{scheibenreif2023masked}
Linus Scheibenreif, Michael Mommert, and Damian Borth.
\newblock Masked vision transformers for hyperspectral image classification.
\newblock In \emph{Proceedings of the IEEE/CVF conference on computer vision
  and pattern recognition}, pages 2166--2176, 2023.

\bibitem[Shen et~al.(2021)Shen, Yang, Wei, Deng, Huang, Hua, Cheng, and
  Liang]{shen2021dct}
Xing Shen, Jirui Yang, Chunbo Wei, Bing Deng, Jianqiang Huang, Xian-Sheng Hua,
  Xiaoliang Cheng, and Kewei Liang.
\newblock Dct-mask: Discrete cosine transform mask representation for instance
  segmentation.
\newblock In \emph{Proceedings of the IEEE/CVF conference on computer vision
  and pattern recognition}, pages 8720--8729, 2021.

\bibitem[Sun et~al.(2022)Sun, Zhao, Zheng, and Wu]{sun2022spectral}
Liang Sun, Guangming Zhao, Yisong Zheng, and Zuxin Wu.
\newblock Spectral--spatial feature tokenization transformer for hyperspectral
  image classification.
\newblock \emph{IEEE Transactions on Geoscience and Remote Sensing},
  60:\penalty0 1--14, 2022.

\bibitem[Theiler et~al.(2019)Theiler, Ziemann, Matteoli, and
  Diani]{theiler2019spectral}
James Theiler, Amanda Ziemann, Stefania Matteoli, and Marco Diani.
\newblock Spectral variability of remotely sensed target materials: Causes,
  models, and strategies for mitigation and robust exploitation.
\newblock \emph{IEEE Geoscience and Remote Sensing Magazine}, 7\penalty0
  (2):\penalty0 8--30, 2019.

\bibitem[Ulicny et~al.(2019)Ulicny, Krylov, and Dahyot]{ulicny2019harmonic}
Matej Ulicny, Vladimir~A Krylov, and Rozenn Dahyot.
\newblock Harmonic networks for image classification.
\newblock 2019.

\bibitem[Ulicny et~al.(2022)Ulicny, Krylov, and Dahyot]{ulicny2022harmonic}
Matej Ulicny, Vladimir~A Krylov, and Rozenn Dahyot.
\newblock Harmonic convolutional networks based on discrete cosine transform.
\newblock \emph{Pattern Recognition}, 129:\penalty0 108707, 2022.

\bibitem[Wang et~al.(2019)Wang, Yong, and Xue]{wang2019frequency}
Ke Wang, Bin Yong, and Zhaohui Xue.
\newblock Frequency domain-based features for hyperspectral image
  classification.
\newblock \emph{IEEE Geoscience and Remote Sensing Letters}, 16\penalty0
  (9):\penalty0 1417--1421, 2019.

\bibitem[Xu et~al.(2023)Xu, Zhao, and Fu]{Xu2023grsl}
Jingran Xu, Jiankang Zhao, and Yuchao Fu.
\newblock An efficient hyperspectral image classification method using deep
  fusion of {3D} discrete wavelet transform and {CNN}.
\newblock \emph{IEEE Geoscience and Remote Sensing Letters}, 20:\penalty0 1--5,
  2023.

\bibitem[Xu et~al.(2020)Xu, Qin, Sun, Wang, Chen, and Ren]{xu2020learning}
Kai Xu, Minghai Qin, Fei Sun, Yuhao Wang, Yen-Kuang Chen, and Fengbo Ren.
\newblock Learning in the frequency domain.
\newblock In \emph{Proceedings of the IEEE/CVF conference on computer vision
  and pattern recognition}, pages 1740--1749, 2020.

\bibitem[Yan et~al.(2024)Yan, Wang, Zhu, and Huang]{yan2024exploiting}
Muge Yan, Lizhi Wang, Lin Zhu, and Hua Huang.
\newblock Exploiting frequency correlation for hyperspectral image
  reconstruction.
\newblock \emph{arXiv preprint arXiv:2406.00683}, 2024.

\bibitem[Yang et~al.(2018)Yang, Ye, Li, Lau, Zhang, and
  Huang]{yang2018hyperspectral}
Xiaofei Yang, Yunming Ye, Xutao Li, Raymond~YK Lau, Xiaofeng Zhang, and Xiaohui
  Huang.
\newblock Hyperspectral image classification with deep learning models.
\newblock \emph{IEEE Transactions on Geoscience and Remote Sensing},
  56\penalty0 (9):\penalty0 5408--5423, 2018.

\bibitem[Yang et~al.(2020)Yang, Zhang, Ye, Lau, Lu, Li, and
  Huang]{yang2020synergistic}
Xiaofei Yang, Xin Zhang, Yun Ye, Raymond~Y Lau, Sheng Lu, Xin Li, and Xiaodong
  Huang.
\newblock Synergistic {2D/3D} convolutional neural network for hyperspectral
  image classification.
\newblock \emph{Remote Sensing}, 12\penalty0 (12):\penalty0 2033, 2020.

\bibitem[Yang et~al.(2022)Yang, Cao, Lu, and Zhou]{yang2022hyperspectral}
Xiaofei Yang, Weijia Cao, Yanan Lu, and Yicong Zhou.
\newblock Hyperspectral image transformer classification networks.
\newblock \emph{IEEE Transactions on Geoscience and Remote Sensing},
  60:\penalty0 1--15, 2022.

\bibitem[Yao et~al.(2024)Yao, Hong, Li, and Chanussot]{yao2024spectralmamba}
Jing Yao, Danfeng Hong, Chenyu Li, and Jocelyn Chanussot.
\newblock Spectralmamba: Efficient mamba for hyperspectral image
  classification.
\newblock \emph{arXiv preprint arXiv:2404.08489}, 2024.

\bibitem[Yao et~al.(2022)Yao, Pan, Li, Ngo, and Mei]{yao2022wave}
Ting Yao, Yingwei Pan, Yehao Li, Chong-Wah Ngo, and Tao Mei.
\newblock Wave-vit: Unifying wavelet and transformers for visual representation
  learning.
\newblock In \emph{European Conference on Computer Vision}, pages 328--345.
  Springer, 2022.

\bibitem[Zhang et~al.(2024)Zhang, Zhang, Liu, Shi, and Pan]{zhang2024three}
Jun Zhang, Cheng Zhang, Shuai Liu, Zhenwei Shi, and Bin Pan.
\newblock Three-dimensional frequency domain transform network for cross-scene
  hyperspectral image classification.
\newblock \emph{IEEE Transactions on Geoscience and Remote Sensing}, 2024.

\bibitem[Zhang et~al.(2016)Zhang, Zhang, and Du]{zhang2016deep}
Liangpei Zhang, Lefei Zhang, and Bo Du.
\newblock Deep learning for remote sensing data: A technical tutorial on the
  state of the art.
\newblock \emph{IEEE Geoscience and Remote Sensing Magazine}, 4\penalty0
  (2):\penalty0 22--40, 2016.

\bibitem[Zhou et~al.(2024)Zhou, Kamata, Wang, Wong, and Hou]{zhou2024mamba}
Weilian Zhou, Sei-Ichiro Kamata, Haipeng Wang, Man~Sing Wong, and
  Huiying~Cynthia Hou.
\newblock {Mamba-in-Mamba}: Centralized {Mamba-Cross-Scan} in {Tokenized Mamba}
  model for hyperspectral image classification.
\newblock \emph{Neurocomputing}, page 128751, 2024.

\bibitem[Zhu et~al.(2024)Zhu, Liao, Zhang, Wang, Liu, and Wang]{zhu2024vision}
Lianghui Zhu, Bencheng Liao, Qian Zhang, Xinlong Wang, Wenyu Liu, and Xinggang
  Wang.
\newblock Vision mamba: Efficient visual representation learning with
  bidirectional state space model.
\newblock \emph{arXiv preprint arXiv:2401.09417}, 2024.

\end{thebibliography}
}

\end{document}